\documentclass{article}
\usepackage[letterpaper,top=2cm,bottom=2cm,left=2cm,right=1.8cm,marginparwidth=1.75cm]{geometry}

\usepackage{amsthm}
\newtheoremstyle{boldremark}
  {\topsep}                    
  {\topsep}                    
  {\normalfont}                
  {}                           
  {\bfseries}                  
  {.}                          
  {.5em}                       
  {}                           
\theoremstyle{boldremark}

\usepackage{booktabs}
\usepackage[pdfencoding=auto, psdextra]{hyperref}
\usepackage{todonotes}

\usepackage{hyperref}
\usepackage{amssymb,amsmath,amsthm,bm}

\usepackage{amsthm} 

\usepackage{authblk}
\usepackage{xcolor}
\usepackage{pifont}
\newcommand{\cmark}{\textcolor{green}{\ding{51}}} 
\newcommand{\omark}{\textcolor{orange}{\ding{51}}} 
\newcommand{\xmark}{\textcolor{red}{\ding{55}}}   

\usepackage{enumitem}
\usepackage{float}

\usepackage{amsthm}
\theoremstyle{remark}
\newtheorem*{remark}{\textbf{Remark}}

\usepackage{graphicx}
\usepackage{subcaption}
\usepackage{float}
\usepackage{multirow}
\usepackage{rotating}
\usepackage{pdflscape}

\usepackage{algorithm}
\usepackage{algorithmic}
\usepackage{amsmath}

\usepackage{microtype}
\usepackage{comment}
\usepackage{soul}
\usepackage{todonotes}

\usepackage{hyperref}
\hypersetup{
    colorlinks=true,
    linkcolor=blue,
    citecolor=blue,
    urlcolor=blue
}

\title{TB-Transolver: Parallelized and Coherent Long-Horizon Rollouts via Temporal Bundling for Time-Dependent PDEs}

\author[]{Sidharth S. Menon}
\author[]{Ameya D. Jagtap\thanks{Corresponding author: Ameya D. Jagtap (ajagtap@wpi.edu, ameyadjagtap@gmail.com)}}

\affil[]{\textit{\small{Aerospace Engineering Department, Worcester Polytechnic Institute, Worcester, MA 01609, USA.}}}
\date{}
\begin{document}

\maketitle

\begin{abstract}
\noindent Transolver has recently gained widespread adoption for solving diverse physical problems on unstructured geometric meshes by leveraging its Physics-Attention mechanism, which seamlessly extends standard softmax attention to handle physical domains at scale. For time-dependent problems, existing literature relies on time-conditioned or autoregressive rollouts to preserve temporal causality while retaining Physics-Attention as the spatial backbone. However, a fundamental limitation of this formulation is its sequential execution model, wherein computational costs scale linearly with the temporal horizon, rendering standard Transolver inefficient for transient problems despite its exceptional spatial efficacy. This bottleneck has historically hindered its broader application in time-dependent physical modeling. To address this limitation, we introduce \textbf{TB-Transolver} (also referred to as Temporally Bundled Transolver), an efficient framework built on temporal bundling that scales Transolver for time-dependent PDEs. Furthermore, we propose \emph{spatiotemporal coherence} (${\rm C(K)}$) as a general metric to systematically quantify snapshot correlation within a temporal bundle, enabling the \emph{a priori} selection of optimal bundle widths using training data alone, thereby avoiding expensive grid searches. We demonstrate the computational efficacy of TB-Transolver across four physically distinct domains, namely, plasticity, fluid dynamics, reaction--diffusion, and global weather forecasting; achieving training speedups of up to $29\times$ over conventional autoregressive and time-conditioned baselines. Finally, these computational gains are further amplified through distributed multi-node and multi-GPU parallelism, yielding aggregated speedups of up to $45.6\times$ at optimal bundle widths for systems exhibiting high spatiotemporal coherence compared to single-GPU implementations.
\end{abstract}

\maketitle
\vspace{0.2cm}
\noindent
\begin{small}Keywords: \textit{Transolver}, \textit{Temporal Bundling}, \textit{Spatiotemporal Coherence}, \textit{Time-Dependent PDEs}, \textit{Parallel Computing}.
\end{small}

\section{Introduction}

In the era of physical AI, leveraging advanced deep learning architectures to solve the partial differential equations (PDEs) governing physical systems has gained significant momentum alongside conventional numerical methods, such as the finite element and finite volume methods. While early efforts primarily focused on solving individual PDEs formulated as specific learning objectives, recent research has increasingly shifted toward learning general operators that capture input--output mappings across a family of related physical systems. Although established methods such as the Fourier Neural Operator (FNO) \cite{li2020fourier} demonstrate scalability through operations in the Fourier domain, they often face limitations when handling complex, unstructured spatial domains commonly encountered in real-world applications. To address these challenges, several recent approaches have focused on efficiently representing high-dimensional spatial information by projecting irregular meshes into learned latent spaces. Other methods incorporate graph-based operations \cite{li2020neural}, geometric Fourier transforms \cite{li2023geometry}, or U-Net architectures \cite{rahman2022u} to capture multiscale spatial correlations in physical systems. However, graph-based kernels and the periodicity assumptions underlying Fourier bases can limit the ability of these methods to represent complex geometries and facilitate global information transfer.

A parallel line of research in neural PDE solvers has been inspired by recent advances in large language models (LLMs), particularly the Transformer architecture \cite{vaswani2017attention}, which introduced the attention mechanism for modeling sequential text data. A recent comprehensive survey of scientific foundation models (SciFMs) \cite{menon2025scientific} reports that more than $80\%$ of SciFM architectures across a broad range of applications, including PDE solving, are based on Transformers. Despite their widespread adoption, Transformer-based architectures face an inherent computational bottleneck arising from the quadratic complexity of the softmax attention mechanism. To mitigate this limitation, neural PDE solvers such as Fact-Former \cite{li2023scalable} employ factorized attention, while GNOT \cite{hao2023gnot}, OFormer \cite{li2022transformer}, and related approaches leverage linear attention mechanisms to reduce computational complexity. However, these modifications often rely on mathematical approximations that can compromise attention quality and, in some cases, degrade performance on general-purpose tasks.

Furthermore, existing architectures exhibit limitations in handling unstructured meshes---some apply attention directly to large numbers of mesh points, incurring substantial computational costs, while others are restricted to structured meshes. In contrast, Transolver \cite{wu2024transolver} efficiently processes spatially unstructured meshes while retaining the standard softmax attention mechanism without mathematical approximations. Its efficiency stems from applying attention to intrinsic physical states, represented by learnable intermediate spatial slices through a mechanism termed as Physics-Attention, rather than directly to individual spatial grid points. Despite its demonstrated applicability to industrial-scale problems, including automotive crash dynamics \cite{nabian2026automotive} and steady-state external aerodynamics \cite{zhou2026transolver}, Transolver has been identified as facing scalability challenges when solving time-dependent PDEs. As highlighted by a recent study \cite{wen2025gaot}---``\textit{In fact, models like Transolver and GNOT scale very poorly, making it impossible for us to train them for the large-scale time-dependent datasets....}".

Recent efforts have extended Transolver's spatial scalability to industry-scale meshes containing hundreds of millions of elements through distributed spatial computation across multiple GPUs in Transolver++ \cite{luo2025transolver++} and improvements to the computational efficiency of slice/deslice operations in Transolver-3 \cite{zhou2026transolver}. However, these advances primarily target spatial scalability and do not directly address the architecture's computational bottlenecks along the temporal dimension. This gap reflects the broader emphasis of existing research on improving spatial attention mechanisms and developing efficient representations of complex, unstructured meshes, with comparatively less ``attention" devoted to scaling neural PDE solvers for time-dependent problems.

Another recent work \cite{hu2026linearno} argues that Physics-Attention is redundant, attributing much of Transolver's effectiveness to the slice/deslice operations rather than explicit interactions among physical slices. In contrast, a subsequent work \cite{shangguan2026trans-sigma} introduces a residual refinement mechanism that retains Physics-Attention to capture non-local interactions among physical slices while preserving the underlying physical representations. In the present work, we retain the original Physics-Attention \cite{wu2024transolver} mechanism, while noting that our temporal-bundling strategy is, in principle, applicable to alternative architectures such as LinearNO \cite{hu2026linearno} and Transolver-$\sigma$ \cite{shangguan2026trans-sigma}.\\

 \noindent \textbf{Autoregressive neural PDE solvers}, illustrated in Figure \ref{fig:tb-transolver} as sequential discrete-time computation, accumulate prediction errors at each rollout step, a phenomenon widely recognized as the \emph{distribution shift problem}~\cite{brandstetter2022message}. In this setting, small single-step errors can compound over time, driving predictions away from the training data distribution and potentially inducing severe instabilities during long-horizon rollouts. For time-dependent PDEs, autoregressive evolution advances the solution to the next time instance according to
\begin{align}
    \label{eq:autoregressive}
    u(t + \Delta t) = \mathcal{A}(\Delta t, u(t)),
\end{align}

where $\mathcal{A}: \mathbb{R}^+ \times \mathbb{R}^N \to \mathbb{R}^N$ denotes the autoregressive operator, which maps the current state to a future state after a time increment $\Delta t$. In contrast, time-conditioned formulations predict future states directly from the target timestamp, without explicitly conditioning on previously predicted states:
\begin{align}
    \label{eq:timeconditioned}
    u(t + \Delta t) = \mathcal{T}(t+\Delta t, \mathcal{I}),
\end{align}

\noindent where $\mathcal{T}: \mathcal{I} \times \mathbb{R}^+ \to \mathbb{R}^N$ denotes the time-conditioned operator, which evaluates the state vector at a given timestamp, conditioned on the problem-specific information $\mathcal{I}$. The latter may include, for example, the die shape in the plasticity benchmark. Thus, while autoregressive models typically advance chronologically using a fixed time step $\Delta t$ and previously computed states, time-conditioned models query the solution directly at the desired time. Although both paradigms are widely used in the literature, autoregressive formulations remain prevalent in neural operator learning because they explicitly propagate predictions through successive time steps. However, this sequential dependence also exposes them to error accumulation and distribution shift. Investigating this vulnerability, McCabe \textit{et al.}~\cite{mccabe2023towards} show that nonlinear operations in FNO can generate unresolved high-frequency modes that alias into lower frequencies, contributing to error accumulation and instability during autoregressive rollouts.


Temporal bundling reduces the number of model evaluations required to reach a given time horizon by predicting $\rm K$ temporal snapshots within a single forward pass, spanning a temporal interval of $\rm (K-1)\Delta t$, thereby reducing the number of model evaluations required to reach a given time horizon and limiting error accumulation during rollouts. This strategy has since been adopted across a range of neural operator architectures, including DeepONet variants~\cite{diab2025tno} and Transformer-based operators~\cite{ovadia2023ditto}, as well as domain-specific surrogates for photonics~\cite{ma2024pic2osim}, weather forecasting~\cite{levang2026gem3}, and electronic structure~\cite{zhang2026orbevo}. Despite its widespread adoption, the temporal bundle width, $\rm K$, is typically fixed by convention or selected through ablation studies~\cite{diab2025tno}, which can be computationally expensive. Moreover, existing studies report varying observations regarding the influence of temporal bundling on autoregressive model training. For instance, small bundle widths can lead to overfitting to the one-step prediction map and divergence during long-horizon rollouts, whereas excessively large bundles can degrade predictive accuracy~\cite{zhang2026orbevo}. Other studies report error spikes at bundle boundaries~\cite{gopakumar2024plasma} and suggest that larger bundles increase computational costs and compromise temporal causality~\cite{ma2024pic2osim}, potentially weakening correlations between successive output frames. \emph{Despite these observations, existing studies do not provide a quantitative criterion for selecting $\rm K$ before training, leaving temporal bundle-width selection for time-dependent problems largely reliant on costly empirical tuning.}

A related line of research focuses on selecting the autoregressive \emph{stride}, $\rm \Delta t$, rather than the temporal bundle width. In three-dimensional turbulence, a recent study \cite{zou2026uq} demonstrates that the temporal autocorrelation function (ACF) of the flow field is correlated with neural-operator stability. Specifically, excessively large time intervals weaken correlations between successive snapshots, whereas excessively small intervals introduce redundancy and prolong autoregressive rollouts. These observations have further motivated the development of multi-stride architectures \cite{pan2026msmoe}. Similar trade-offs have been reported in studies of step-size selection for neural PDE solvers \cite{lippe2023pderefiner} and weather emulators \cite{levang2026gem3}. More recently, Chen \textit{et al}. \cite{chen2026flux} proposed selecting the \emph{input} history length based on the decorrelation time of a closure diagnostic. However, these criteria rely on temporal correlations evaluated at fixed spatial locations and target either the autoregressive stride, $\rm \Delta t$, or the input history length, rather than the temporal bundle width, $\rm K$. 

A complementary direction involves learning the rollout step size during training. TANTE \cite{wu2025tante}, for instance, jointly learns higher-order temporal derivatives and a local radius of convergence through a neural Taylor expansion, allowing the rollout step size to adapt to local dynamics. Although such approaches enable trajectory-dependent adaptation, they require additional training to learn the step-size selection mechanism and introduce architectural complexity. \emph{Consequently, selecting the temporal bundle width before training remains an open challenge.}

\noindent The proposed work bridges the gap between these lines of research. We introduce \emph{spatiotemporal coherence}, a training-free, data-derived quantity that measures spatial correlation across the span of a candidate temporal bundle. We empirically demonstrate its utility for selecting a suitable bundle width $\rm K$ \emph{a priori}, before training. Unlike temporal-only ACF-based criteria~\cite{zou2026uq}, the proposed metric quantifies spatial-field correlation across the temporal span of a candidate bundle, providing a complementary perspective for assessing spatiotemporal coherence. Unlike learned adaptive schemes~\cite{wu2025tante}, it incurs no training-time overhead and does not modify or complicate the architecture. Building on temporal bundling, we demonstrate substantial speedups when training Transolver on time-dependent PDEs across a diverse set of physical benchmarks. Our main contributions are as follows:
\begin{itemize}
    \item We identify incremental trajectory resolution---advancing one time step at a time---as a primary computational bottleneck limiting Transolver on time-dependent PDEs and address it through temporal bundling. Termed TB-Transolver, this approach achieves training speedups of up to $29\times$ and substantially reduces memory consumption while preserving predictive accuracy. These improvements enable successful training on datasets and configurations that exceed the available GPU memory capacity of the baseline Transolver~\cite{wu2024transolver}.
    \item We introduce \emph{spatiotemporal coherence} (${\rm C(K)}$), a training-free, data-driven measure of snapshot correlation within a candidate temporal bundle, and use it to systematically select the bundle width ${\rm K}$ prior to training. We demonstrate that severe decorrelation among snapshots within a bundle characterizes and helps explain the practical limits of temporal bundling across diverse physical dynamics.
    \item While temporal bundling reduces computational cost by predicting multiple time steps per forward pass, multi-GPU scaling reduces wall-clock time by distributing spatial computations across multiple GPUs. Combining temporal bundling with multi-GPU scaling yields aggregate speedups of up to $45.6\times$ for time-dependent PDE problems modeled using TB-Transolver.
    \item We comprehensively evaluate TB-Transolver across four physically distinct domains—fluid dynamics, plasticity, weather forecasting, and reaction--diffusion—demonstrating its effectiveness for both autoregressive and time-conditioned temporal rollout formulations.
\end{itemize}

\section{Methodology}

\begin{figure}[H]
    \centering
    \includegraphics[width=\linewidth, trim={0cm 3.25cm 0cm 2.5cm}, clip]{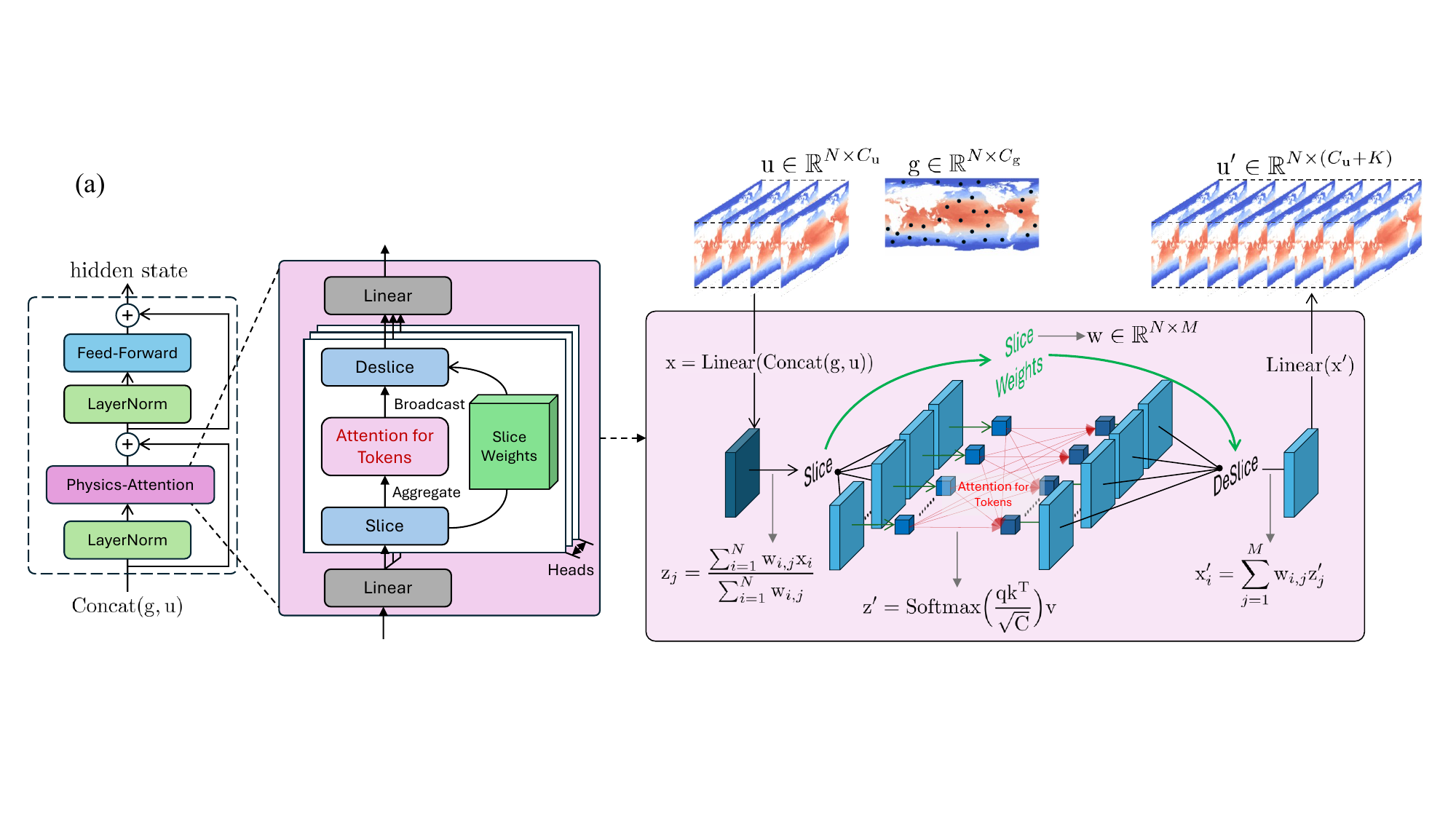}

    \vspace{0.5em}

    \includegraphics[width=\linewidth, trim={0.5cm 1.5cm 0.5cm 2cm}, clip]{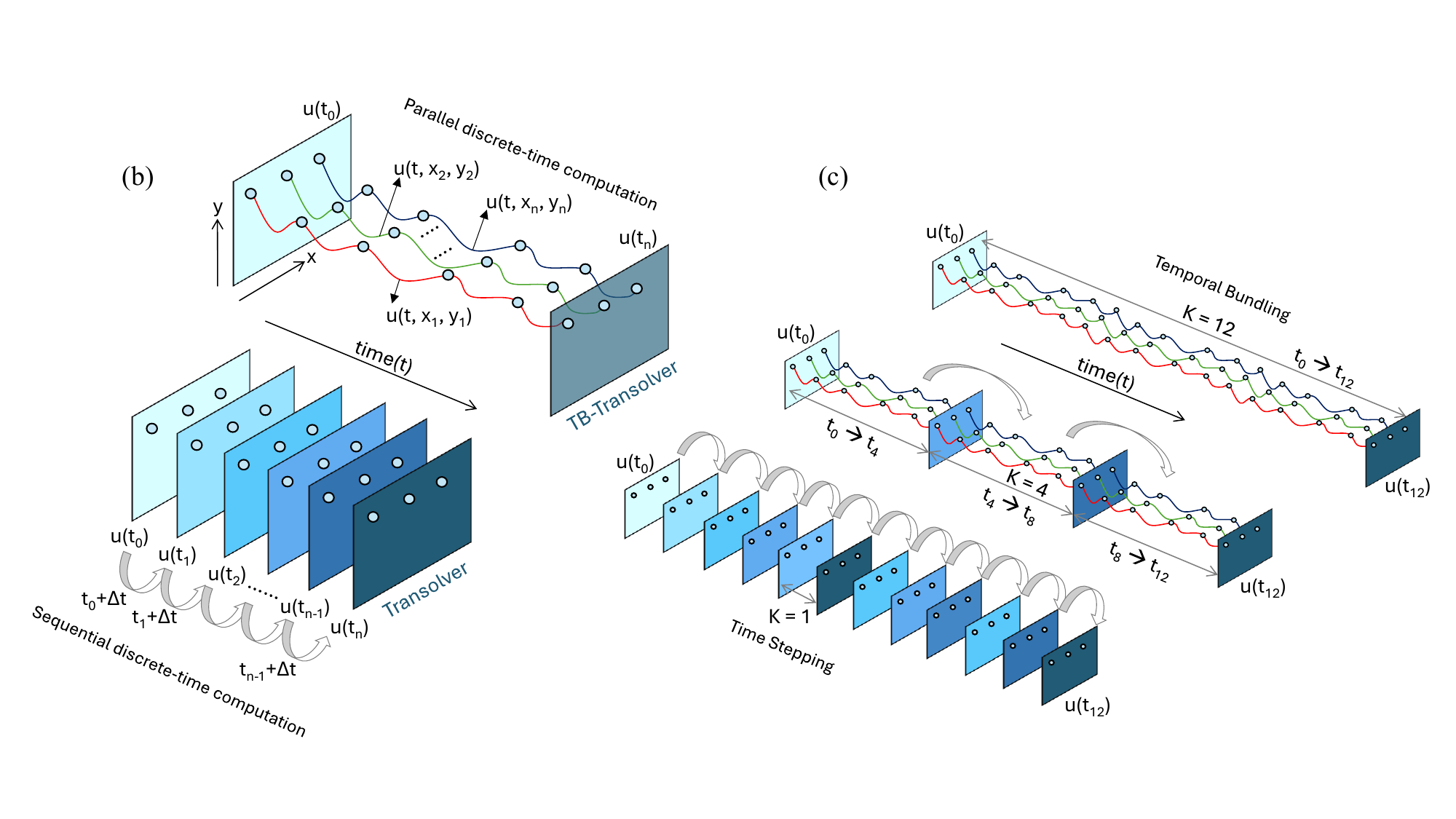}

    \caption{(a) Overview of the spatial physics-aware interactions and temporal parallelism in TB-Transolver. The Physics-Attention mechanism aggregates spatial information into physics-aware slices, which serve as macro-tokens for capturing physical correlations across the spatial domain. Attention is then applied to these tokens to model interactions among the physical slices. (b) Comparison of sequential and parallel discrete-time computations. Sequential computation involves \textbf{time-conditioned} or \textbf{autoregressive rollouts}, whereas \textbf{temporal bundling} enables multiple time steps to be predicted in parallel. (c) In TB-Transolver, autoregressive time stepping corresponds to the special case of temporal bundling with bundle width $\rm K=1$, whereas $\rm K>1$ enables parallel prediction across multiple time steps. The illustration shows an example with 12 temporal snapshots.}
    \label{fig:tb-transolver}
\end{figure}

\textbf{Transolver}: The backbone of Transolver \cite{wu2024transolver} that facilitates computational efficiency while handling large unstructured meshes is the Physics-Attention mechanism as shown in Figure \ref{fig:tb-transolver}. Unlike the original softmax attention mechanism in Transformer \cite{vaswani2017attention}, the fundamental hypothesis of the Physics-Attention mechanism is applying softmax attention over an aggregated $M$ physics-aware macro-tokens instead of the $N$ spatial points from the unstructured mesh. This leads to an overall computational complexity of $\mathcal{O}(NMC + M^2C)$ which is approximately linear when $M << N$. Here, $C$ denotes the number of channels of the deep input feature, $\mathrm{x} \in \mathbb{R}^{N\times C}$ linearly projected from the concatenated input geometry, $\mathrm{\textbf{g}} \in \mathbb{R}^{N \times C_{\mathrm{g}}}$ and initial observations, $\mathrm{\textbf{u}} \in \mathbb{R}^{N \times C_{\mathrm{u}}}$. It is noteworthy that, $\mathrm{\textbf{u}}$ represents history of past states for autoregressive time rollouts while it represents time instances for time-conditioned rollouts. 

\noindent The projected physics-aware $M$ slices, $\mathrm{\textbf{z}} = \{\mathrm{\textbf{z}}_j\}_{j=1}^M \in \mathbb{R}^{M \times C}$ transitions the Physics-Attention to capture the intricate spatial correlations from the latent states as follows:

\begin{align*}
\mathrm{\textbf{q}}, \mathrm{\textbf{k}}, \mathrm{\textbf{v}} = \mathrm{Linear}(\mathrm{\textbf{z}}),~\mathrm{\textbf{z}}^{\prime} = \mathrm{Softmax}\biggl(\frac{\mathrm{\textbf{q}}\mathrm{\textbf{k}}^\mathrm{T}}{\sqrt{C}}\biggr)\mathrm{\textbf{v}}.     
\end{align*}

\noindent Here, $\mathrm{\textbf{q}},~\mathrm{\textbf{k}},~\mathrm{\textbf{v}},~\mathrm{\textbf{z}}^{\prime}\in \mathbb{R}^{M \times C}$ and $\mathrm{\textbf{z}}^{\prime}$ is projected back to the original mesh points through deslicing which is essentially broadcasting back using the same slice weights, $\mathrm{\textbf{w}} \in \mathbb{R}^{N\times M}$ which were previously used for slicing the input spatial geometry. Despite the linear computational complexity of the Physics-Attention mechanism, their utilization to time-dependent problems is fairly limited due to the inefficiencies while handling the time dimension. Furthermore, time dimension can only be handled sequentially even when the compute hardware such as the GPUs allow parallelization across spatial dimensions. To this end, Transolver serially computes time rollouts with sequential forward passes at small time steps leading to a major computational bottleneck preventing their scalability to time-dependent PDEs of practical relevance.\\

\noindent\textbf{Spatiotemporal Coherence}: In the current work, we observe that autoregressive frameworks must strike a balance between two sources of error: accumulation error ($\varepsilon_{\rm accumulation}$) and decoherence error ($\varepsilon_{\rm decoherence}$). The relative $\rm L_2$ error can be viewed as a cumulative measure of these errors in the context of autoregressive rollouts. While accumulation error becomes more pronounced over long-horizon rollouts, decoherence error depends on the underlying physical dynamics and the sampling time step $\Delta \rm t_{\rm data}$ of the dataset. For instance, in fluid dynamics, if $\Delta \rm t_{\rm data} \gg \tau_{\rm diffusion}, \tau_{\rm convection}$, where $\tau_{\rm diffusion}$ and $\tau_{\rm convection}$ denote the diffusion and convection timescales, respectively, temporal bundling with an arbitrary width $\rm K$ can result in substantial decorrelation among the output snapshots within a bundle. This can increase $\varepsilon_{\rm decoherence}$, potentially offsetting the reduction in $\varepsilon_{\rm accumulation}$ achieved through temporal bundling. To address this trade-off, we propose a metric to quantify the influence of bundle width $\rm K$ on $\varepsilon_{\rm decoherence}$, enabling more informed bundle-width selection and better control of prediction error.

The current work quantifies the spatial correlation between two snapshots separated by a temporal bundle of width $\rm K$ using a metric termed spatiotemporal coherence ($\rm C(K)$). This metric enables informed decisions, prior to training, about the suitability of temporal bundling for a given physical dynamics using a scalable neural PDE solver such as Transolver. Specifically, $\rm C(K)$ is defined as the spatial Pearson correlation between the initial snapshot at $\rm t_0$ and the terminal snapshot at $\rm t_{K-1}$ of a temporal bundle of width $\rm K$:
\begin{align}
    \label{eq:c_k}
    \boxed{\rm C(K) = \frac{\langle u(x,t_0) - \overline{u}_0, u(x,t_{\rm K-1}) - \overline{u}_{\rm K-1}\rangle}{\|u(x,t_0) - \overline{u}_0\|_2 \|u(x,t_{\rm K-1}) - \overline{u}_{\rm K-1}\|_2}}
\end{align}

\noindent Here, $\rm \overline{u}_0 = \overline{u(x,t_0)}$ and $\rm \overline{u}_{K-1} = \overline{u(x,t_{K-1})}$ denote the spatial averages at the initial and terminal snapshots, respectively. Although prior works employ temporal bundling to reduce potential future-trajectory drift, the bundle width is typically treated as a fixed hyperparameter that must be identified through computationally expensive brute-force training searches. In contrast, the proposed strategy provides a physics-guided, zero-cost diagnostic that can be evaluated entirely from the training data before model training, enabling informed selection of the bundle width $\rm K$. Furthermore, we note that
\begin{align*}
\varepsilon_{\rm accumulation} \propto \frac{1}{\rm K}
\quad \text{and} \quad
\varepsilon_{\rm decoherence} \propto \left(1 - \rm C(K;\Delta t_{\rm data})\right).
\end{align*}

\noindent These proportionalities are intended as qualitative heuristics to describe the competing effects of temporal bundling, rather than quantitative error bounds or theoretically derived scaling laws. Although $\Delta\rm t_{data}$ is not a control variable, it characterizes the temporal resolution of the training dataset and can therefore inform the selection of a suitable bundle width before model training. For benchmarks with multiple output channels, such as plasticity (4 channels) and reaction--diffusion (2 channels), the spatial dimensions of all output channels are flattened into a single vector before computing $\rm C(K)$ using Equation \eqref{eq:c_k}. The correlation is computed independently for each training sample, and the resulting values are averaged across the training dataset.\\

\noindent \textbf{Memory Efficiency}: Beyond the significant wall-clock speedups achieved through temporal bundling, TB-Transolver also improves memory efficiency. By reducing the number of model evaluations required for autoregressive rollouts, temporal bundling shortens the computational graph and reduces peak memory consumption.\\

\noindent\textbf{Influence of Time Bundling on Transolver}: Despite its state-of-the-art performance on complex, unstructured spatial domains, Transolver has seen limited adoption as a baseline for time-dependent problems. For instance, GAOT \cite{wen2025gaot} benchmarks Transolver on steady-state problems but omits it from time-dependent experiments, citing poor training scalability on large time-dependent datasets.

\noindent The original Transolver work~\cite{wu2024transolver} addresses time-dependent PDEs using two formulations: time conditioning and autoregressive prediction. In the time-conditioned setting, the model receives the input fields and a query time $\rm t$ and predicts the solution field at that instant. In the autoregressive setting, the model receives past solution states and predicts the next state, repeating this process up to a prescribed time horizon. Both formulations share the same structural limitation: each forward pass predicts the solution at only one time instance. Physics-Attention reduces the per-pass computational cost on unstructured domains from quadratic to linear in the number of mesh points $N$, yielding a complexity of approximately $\mathcal{O}(NM)$, where $M$ is the number of learned slices. However, resolving a trajectory spanning $T$ time instances still requires $T$ forward passes, resulting in a total computational cost of $\mathcal{O}(NMT)$. In the autoregressive setting, these passes must also be executed sequentially, allowing prediction errors to accumulate over time. \emph{Thus, for long-horizon predictions, the spatial efficiency gained through Physics-Attention is offset by the cost of resolving the solution one time instance at a time.} We identify this limitation as a key bottleneck to scaling Transolver to time-dependent PDEs and address it systematically through temporal bundling.

In Transolver, both autoregressive and time-conditioned rollouts follow the time-stepping strategy illustrated in Figure~\ref{fig:tb-transolver}. For short-horizon predictions, this strategy supports stable and accurate predictions of future states. However, the need to predict one time instance at a time becomes a bottleneck when scaling the architecture to time-dependent PDEs. To address this limitation, we propose temporal bundling, also illustrated in Figure~\ref{fig:tb-transolver}, with the bundle width selected based on the spatiotemporal coherence ($\rm C(K)$) introduced in the previous section. Figure~\ref{fig:temporal_correlation} presents the spatiotemporal coherence trends, computed before training Transolver, across diverse datasets spanning plasticity, fluid dynamics, reaction--diffusion, and global weather forecasting. Furthermore, Table~\ref{tab:coherence_regimes} provides a set of decision rules to guide the selection of temporal bundling prior to training. Although the present study focuses on addressing this bottleneck in Transolver, the proposed approach can also be extended to other neural PDE solvers.

\subsection{Datasets}
\begin{figure}[H]
    \centering
    \includegraphics[width=0.65\linewidth]{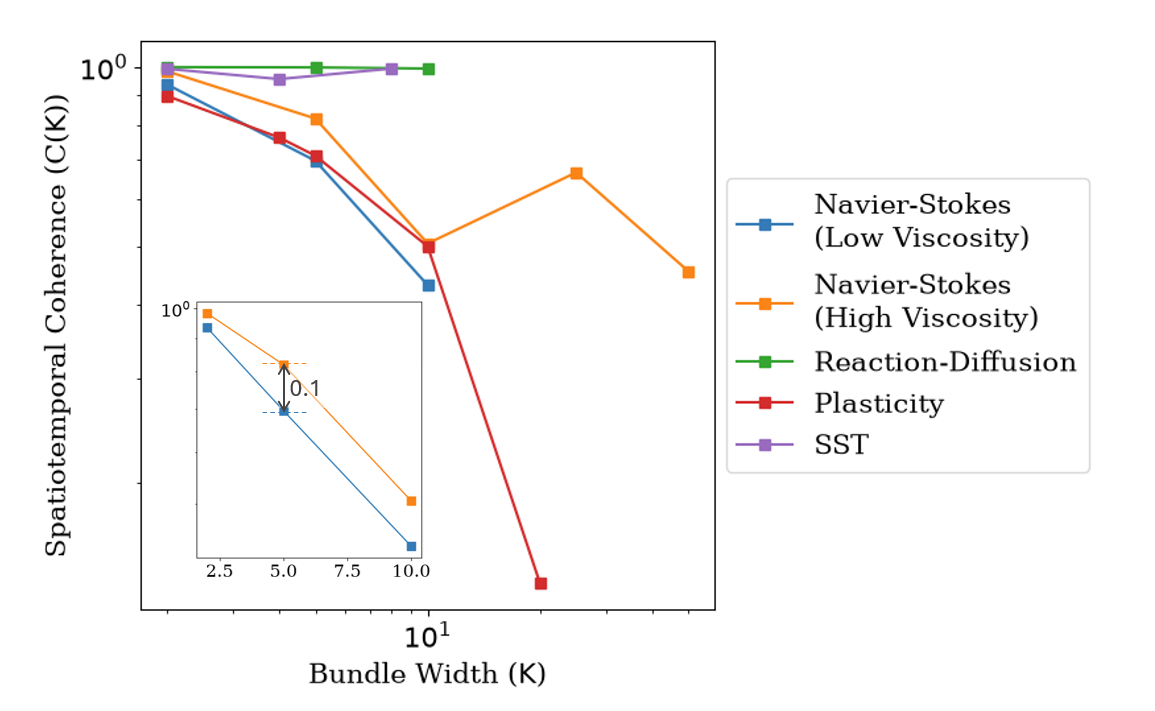}
    \caption{Spatiotemporal coherence (\(C(\rm K)\)) for dynamical systems: (a) low-viscosity Navier–Stokes equation, \(\nu=1\mathrm{e}{-5}\); (b) high-viscosity Navier–Stokes equation, \(\nu=1\mathrm{e}{-3}\); (c) reaction–diffusion equation; (d) plasticity; and (e) sea surface temperature (SST). The curves show the degree of correlation (or decorrelation) as a function of bundle width (\(\rm K\)). When \(C(\rm K)\rightarrow 1\), the first and last slices of a bundle of width \(\rm K\) are highly correlated. Conversely, when \(C(\rm K)\rightarrow 0\), these slices exhibit low correlation, indicating reduced temporal coherence and potentially making temporal bundling less favorable.
    }
    \label{fig:temporal_correlation}
\end{figure}

\noindent \textbf{Plasticity}: To evaluate performance under non-linear elastoplastic solid mechanics, we utilize the 2D plasticity benchmark introduced by Li \textit{et al}.~\cite{li2023geometry}. The task models the deformation dynamics of an elastoplastic material subjected to indentation by arbitrary die geometries. The spatial domain is discretized on a structured grid of resolution ${\rm 101 \times 31}$ (${\rm N = 3,131}$ spatial nodes). The target trajectory spans ${\rm T_{out} = 20}$ future time steps with a uniform temporal sampling increment of ${\rm \Delta t_{data} = 1.0}$, yielding an output field tensor of shape ${\rm 20 \times 101 \times 31 \times 4}$, where the feature channels capture non-dimensional displacements and internal stress field components. To investigate the effect of temporal bundling, we evaluate bundle widths ${\rm K \in \{1, 2, 4, 5, 10, 20\}}$, mapping the full prediction horizon into autoregressive rollouts of length ${\rm T_{out}/K}$. Evaluating the spatiotemporal coherence ${\rm C(K)}$ via Equation~\eqref{eq:c_k} demonstrates a continuous decay (shown in Figure \ref{fig:temporal_correlation}) in spatiotemporal coherence, $\rm C(K)$ as ${\rm K \to T_{out}}$.\\

\noindent \textbf{Fluid Dynamics}: To evaluate performance on fluid flows, we utilize the 2D incompressible Navier--Stokes benchmark in vorticity formulation on a 2D unit torus $\mathbb{T}^2$, introduced by Li \textit{et al}.~\cite{li2020fourier}. The dataset is divided into low-viscosity and high-viscosity regimes to test model stability under varying degrees of spatial turbulence and temporal chaos:
\begin{itemize}
    \item \textbf{Navier--Stokes (Low Viscosity, $\nu = 10^{-5}$):} This regime represents highly chaotic, turbulent flow dynamics. The spatial domain is discretized on a grid resolution of ${\rm 64 \times 64}$ (${\rm N = 4,096}$ spatial nodes). The target trajectory spans an output time horizon of ${\rm T_{out} = 10}$ future time steps with a temporal sampling increment of ${\rm \Delta t_{data} = 1.0}$, given an input sequence of ${\rm T_{in} = 10}$, yielding a target solution tensor of shape ${\rm 10 \times 64 \times 64}$. To investigate temporal bundling, we evaluate bundle widths ${\rm K \in \{1, 2, 5, 10\}}$. Computing the spatiotemporal coherence ${\rm C(K)}$ via Equation~\eqref{eq:c_k} demonstrates a continuous decay (shown in Figure \ref{fig:temporal_correlation}) in spatiotemporal coherence as ${\rm K \to T_{out}}$.
    \item \textbf{Navier--Stokes (High Viscosity, $\nu = 10^{-3}$):} This regime models smooth, strongly dissipative fluid dynamics. The solution trajectory is discretized at a resolution of ${\rm 64 \times 64}$ (${\rm N = 4,096}$). Given an input horizon of ${\rm T_{in} = 20}$, the model predicts an output horizon of ${\rm T_{out} = 30}$ time steps, yielding a target tensor of shape ${\rm 30 \times 64 \times 64}$. We evaluate bundle widths ${\rm K \in \{1, 2, 5, 10, 15, 30\}}$. Computing the spatiotemporal coherence ${\rm C(K)}$ via Equation~\eqref{eq:c_k} reveals a non-monotonic trend (shown in Figure \ref{fig:temporal_correlation}) characterized by initial decay followed by structural oscillations as ${\rm K \to T_{out}}$.
\end{itemize}

\noindent \textbf{Reaction--Diffusion}: To evaluate non-linear pattern formation and coupled dynamics, we utilize the 2D FitzHugh--Nagumo reaction--diffusion system from PDEBench introduced by Takamoto \textit{et al}.~\cite{takamoto2022pdebench}. The benchmark models two non-linearly coupled field variables $({u, v})$ under diffusion coefficients ${D_u = 10^{-3}}$ and ${ D_v = 5 \times 10^{-3}}$. The spatial domain is discretized on a uniform grid of resolution ${\rm 128 \times 128}$ (${\rm N = 16,384}$ spatial nodes). Out of the 100 available temporal snapshots generated with a sampling increment of ${\rm \Delta t_{data} = 0.05}$, we evaluate a sub-sequence of 20 consecutive time steps split into an input horizon of ${\rm T_{in} = 10}$ and a target output prediction horizon of ${\rm T_{out} = 10}$, yielding a target solution tensor of shape ${\rm 10 \times 128 \times 128 \times 2}$. To examine temporal bundling, we evaluate bundle widths ${\rm K \in \{1, 2, 5, 10\}}$. Computing the spatiotemporal coherence ${\rm C(K)}$ via Equation~\eqref{eq:c_k} reveals a predominantly flat trend (shown in Figure \ref{fig:temporal_correlation}) as ${\rm K \to T_{out}}$, indicating that structural coherence is substantially preserved across all admissible bundle widths for this system.\\

\noindent \textbf{Sea Surface Temperature Benchmark}: To evaluate long-range climate forecasting on irregular spatial masks, we utilize hourly single-level Sea Surface Temperature (SST) reanalysis data from ERA5 introduced by Hersbach \textit{et al}.~\cite{hersbach2025era5}, spanning the years 1980 to 2025. The task predicts monthly SST fields recorded on the $15^{\rm th}$ day of each month, where land nodes are masked out since valid physical features are restricted to oceanic coordinates. The temporal sampling increment is ${\rm \Delta t_{data} = 1\text{~month}}$. Given an input history horizon of ${\rm T_{in} = 4\text{~months}}$, the model predicts an output horizon of ${\rm T_{out} = 8\text{~months}}$, yielding an output target tensor of shape ${\rm 8 \times N_{ocean}}$, where ${\rm N_{ocean}}$ denotes the number of active ocean grid points. To evaluate temporal bundling, we select bundle widths ${\rm K \in \{1, 2, 4, 8\}}$. Computing the spatiotemporal coherence ${\rm C(K)}$ via Equation~\eqref{eq:c_k} exhibits a flat trend (shown in Figure \ref{fig:temporal_correlation}) as ${\rm K \to T_{out}}$, demonstrating that structural coherence remains invariant across all candidate bundle widths for this global-scale SST field.\\

\noindent It is worthwhile to emphasize that the candidate bundle widths ${\rm K}$ across all benchmark systems were selected strictly based on exact divisibility (${\rm T_{out}/K}$ integer steps). This choice is adopted for computational convenience to ensure a uniform temporal partition across the forecast horizon ${\rm T_{out}}$. Based on the observed trajectory of the spatiotemporal coherence metric ${\rm C(K)}$ across distinct physical systems, we observe that selecting an optimal bundle width ${\rm K^*}$—which balances computational efficiency with predictive accuracy—is strongly governed by the underlying dynamical properties of the target dataset. We synthesize these general empirical and theoretical observations in Table~\ref{tab:coherence_regimes}. In the subsequent sections, we demonstrate the concrete computational performance gains achieved by temporal bundling for ${\rm K > 1}$ for Transolver, illustrating how an optimal ${\rm K}$ can be systematically identified by considering spatiotemporal coherence alongside the length of the temporal rollout horizon.

\begin{table}[H]
\centering
\small
\begin{tabular}{cccc}
\toprule
\textbf{$\rm C(K)$ Trend} & \textbf{Higher accuracy} & \textbf{Higher efficiency} & \textbf{Source of Error} \\
\midrule
Flat (short/long rollouts)
& when $\rm K \rightarrow T_{out}$
& \multirow{3}{*}{when $\rm K \rightarrow T_{out}$}
& $\varepsilon_{\rm accumulation}$ \\
\cmidrule(lr){1-2}\cmidrule(lr){4-4}
Decays (short rollouts)
& when $\rm K \rightarrow 1$
&
& $\varepsilon_{\rm accumulation},~\varepsilon_{\rm decoherence}~(\uparrow)$ \\
\cmidrule(lr){1-2}\cmidrule(lr){4-4}
Decays (long rollouts)
& at optimal $\rm K^*$ ($\rm 1 < K^* < T_{out}$)
&
& $\varepsilon_{\rm accumulation}~(\uparrow),~\varepsilon_{\rm decoherence}$ \\
\bottomrule
\end{tabular}
\caption{Performance trade-offs based on the structural coherence metric $\rm C(K)$. Note that $(\uparrow)$ represents the dominating error for time rollouts.}
\label{tab:coherence_regimes}
\end{table}

\section{Results}
This section examines the computational gains of temporal bundling in Transolver for both time-conditioned and autoregressive rollouts across a range of time-dependent problems in the physical sciences. All experiments in the present study are conducted using a single NVIDIA RTX PRO 6000 Blackwell GPU, whereas Wu \textit{et al.} evaluated the original Transolver using an NVIDIA A100 GPU \cite{wu2024transolver}. The training configuration for all benchmarks are tabulated and discussed in Appendix \ref{appx:training_config}.

\begin{table}[H]
\centering
\small
\begin{tabular}{lcc}
\toprule
\multirow{2}{*}{\textbf{Model}} & \multirow{2}{*}{\textbf{Plasticity}} & \textbf{Navier--Stokes} \\
 & & ($\nu = 1\mathrm{e}{-5}$)\\
\midrule
FNO \cite{li2020fourier} & -- & 0.1556 \\
WMT \cite{gupta2021multiwavelet} & 0.0076  & 0.1541 \\
U-FNO \cite{wen2022u} & 0.0039 & 0.2231 \\
geo-FNO \cite{li2023geometry} & 0.0074  & 0.1556 \\
U-NO \cite{rahman2022u} & 0.0034 & 0.1713 \\
F-FNO \cite{tran2021factorized} & 0.0047 & 0.2322 \\
Galerkin Transformer \cite{cao2021choose} & 0.0120 & 0.1401 \\
HT-Net \cite{liu2024mitigating} & 0.0333 & 0.1847 \\
OFormer \cite{li2022transformer} & 0.0017 & 0.1705 \\
GNOT \cite{hao2023gnot} & 0.0336 & 0.1380 \\
FactFormer \cite{li2023scalable} & 0.0312 & 0.1214 \\
LSM \cite{wu2023solving} & 0.0025 & 0.1535 \\
ONO \cite{xiao2023improved} & 0.0048 & 0.1195 \\
\midrule
\textbf{Transolver} \cite{wu2024transolver} & \textbf{0.0012} & \textbf{0.0900} \\
\bottomrule
\end{tabular}
\caption{Comparative performance (Relative $\rm L_2$ Error) of baseline operator learning models and Transolver across standard Plasticity and Navier-Stokes PDE benchmarks by Wu \textit{et al.} \cite{wu2024transolver}. Note that here, bundle width with $\rm K = 1$ corresponds to Transolver. Also, plasticity is trained by time-conditioning while Navier-Stokes is trained by auto-regressive rollouts.}
\label{tab:transolver_baselines}
\end{table}

\noindent Therefore, to verify that prediction accuracy is consistent across GPU types, we compare the relative $\rm L_2$ errors obtained using Transolver with temporal bundle width $\rm K = 1$ on our GPU against the results reported in the original work for two benchmark---plasticity and Navier--Stokes. We then demonstrate the computational benefits of temporal bundling for larger bundle widths, selected based on the spatiotemporal coherence $\rm C(K)$ of each dataset representing a time-dependent problem.

Here, we consider two types of temporal rollouts for the Transolver architecture---\textbf{time-conditioned} (Equation \eqref{eq:timeconditioned}) and \textbf{autoregressive} (Equation \eqref{eq:autoregressive}) rollouts. Specifically, we use time-conditioned rollouts for plasticity and autoregressive rollouts for the remaining time-dependent problems---Navier--Stokes, reaction--diffusion, and global weather forecasting. Although these are widely adopted strategies for operator learning of time-dependent problems, we identify the one-time-step-per-forward-pass constraint as a critical bottleneck to scaling Transolver for such problems. To address this limitation, we investigate the feasibility of temporal bundling based on the trend of $\rm C(K)$ for both short and long-horizon rollouts, as summarized in Table~\ref{tab:coherence_regimes}. We evaluate temporal bundling for Navier--Stokes in both high and low-viscosity regimes. The low-viscosity regime presents physical challenges associated with more complex and chaotic flow dynamics, whereas the high-viscosity regime presents a challenge due to the relatively long autoregressive rollout, in which 30 future time steps are predicted given a 20-timestep input context. We further conclude the discussions with distributed data parallelism for Transolver while solving time-dependent problems using the optimal bundle width $\rm K ^*$, and demonstrate the scaling efficiency and the subsequent speed-up at the optimally chosen bundle width. It is worthwhile to note that here, the optimal bundle width $\rm K^*$ is decided based on Table \ref{tab:coherence_regimes} by weighing the gain in computational efficiency at the expense of accuracy.

The optimality of the bundle width discussed here is empirical and is unrelated to optimality conditions in theoretical mathematics. In practice, given a dataset representing the dynamics of a physical problem, one can compute $\rm C(K)$ for a range of bundle widths to identify the trend in spatiotemporal coherence, as illustrated in Figure~\ref{fig:temporal_correlation}, and select an appropriate bundle width using the decision rules in Table~\ref{tab:coherence_regimes}. \emph{Furthermore, our empirical results across benchmarks indicate that $\rm C(K) \gtrsim 0.8$ is associated with favorable accuracy–efficiency trade-off. We therefore recommend considering this threshold when searching for an optimal bundle width $\rm K^*$ across a range of candidate values.}
\begin{align}
\label{eq:c_k_threshold}
\boxed{\rm K^* = \max \left\{ K \;:\; C(K) \gtrsim 0.8 \right\}}
\end{align}

\noindent It is worth noting that Equation \eqref{eq:c_k_threshold} provides an empirically motivated criterion for selecting the bundle width, which is consistent with the benchmark results across the physical domains considered and the different coherence trends observed in Figure \ref{fig:temporal_correlation}. While the current work focuses on Transolver, further investigation across additional architectures, benchmarks, and physical domains is warranted to refine the threshold and assess the broader applicability and potential universality of the proposed selection criterion.

\subsection{Temporally Bundled Transolver for Scaling Efficiency}

The empirical results across the time-dependent benchmarks considered in this section further demonstrate how the trend in $\rm C(K)$, shown in Figure~\ref{fig:temporal_correlation}, relates to prediction accuracy and computational efficiency, as summarized in Table~\ref{tab:coherence_regimes}. Tables~\ref{tab:plasticity}--\ref{tab:high_viscosity_ns} report the performance of temporally bundled Transolver across plasticity, low-viscosity Navier--Stokes, reaction--diffusion, high-viscosity Navier--Stokes, and global sea surface temperature forecasting. Comparing the mean errors for plasticity and Navier--Stokes in Tables~\ref{tab:plasticity} and \ref{tab:navier_stokes}, respectively, against the baseline results reported in Table~\ref{tab:transolver_baselines} for $\rm K = 1$, we find that our results are consistent with those of the original Transolver work~\cite{wu2024transolver}, despite the differences in GPU configurations.

\begin{figure}[H]
    \centering
    \includegraphics[width=1\linewidth, trim={1cm 7.4cm 1cm 6cm}, clip]{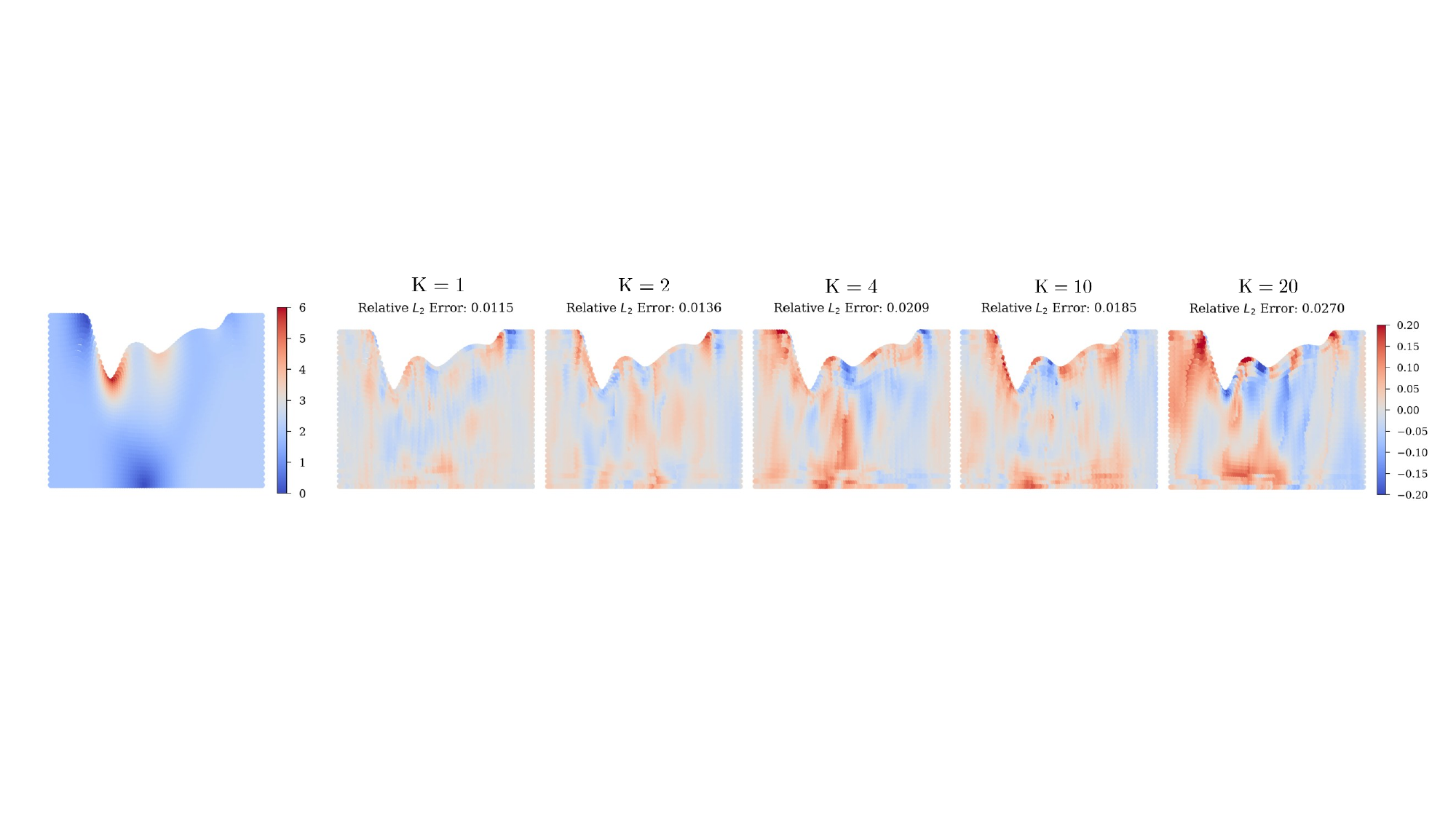}
    \caption{\textbf{Performance of TB-Transolver on the Plasticity Benchmark}: As bundle width ${\rm K}$ increases toward ${\rm T_{out}}$, TB-Transolver achieves a speedup of up to $29\times$ while relative $\rm L_2$ error increases by $125\%$ compared to time-conditioned rollouts at ${\rm K} = 1$.}
    \label{fig:plasticity}
\end{figure}

\begin{table}[H]
\centering
\begin{tabular}{cccccccc}
\toprule
Bundle Width ($\rm K$) & Mean Error & Sec/Epoch & Total Time (500 Ep.) & Peak VRAM & $\rm C(K)$ & Speed-up\\
\midrule
1  & $0.00120^{[1]}$ & 122.00 & 17.0 h & 1.98 GB & 1.0 & $\rm Baseline$\\
2  & $0.00160^{[1]}$ & 59.31  & 8.2 h  & 1.99 GB & 0.894 & $2 \times$\\
4  & $0.00192^{[2]}$ & 30.17  & 4.2 h  & 1.98 GB & 0.761 & $4 \times$\\
5  & $0.00194^{[2]}$ & 22.50  & 3.1 h  & 1.98 GB & 0.709 & $5 \times$\\
10 & $0.00229^{[2]}$ & 6.18   & 52 min & 1.99 GB & 0.499 & $20 \times$\\
20 & $0.00270^{[1]}$ & 4.15   & 35 min & 1.98 GB & 0.135 & $29 \times$\\
\bottomrule
\end{tabular}
\caption{\textbf{Plasticity Benchmark (${\rm T_{out} = 20}$)}: Accuracy and computational efficiency for varying bundle widths. ${\rm C(K)}$ follows the trend illustrated in Figure~\ref{fig:temporal_correlation}. The model is trained using time-conditioned rollouts. The model comprises $2.8$ million parameters.}
\label{tab:plasticity}
\end{table}

As summarized in Table~\ref{tab:coherence_regimes}, computational efficiency generally improves with increasing bundle width $\rm K$, a trend consistently observed across all benchmarks in Tables~\ref{tab:plasticity}--\ref{tab:high_viscosity_ns}. A substantial portion of these gains arises from reducing the number of model evaluations required across the temporal horizon. In the limiting case of $\rm K = T_{out}$, temporal bundling allows the model to predict the entire time horizon in a single forward pass, avoiding the sequential time stepping required by conventional time-conditioned and autoregressive rollouts. Although such one-shot predictions are expected to reduce accumulation error ($\rm \varepsilon_{\rm accumulation}$), this reduction is not guaranteed and depends on the underlying physical dynamics represented by the dataset.\\


\begin{figure}[H]
    \centering
    \includegraphics[width=1\linewidth, trim={2cm 3cm 1cm 1cm}, clip]{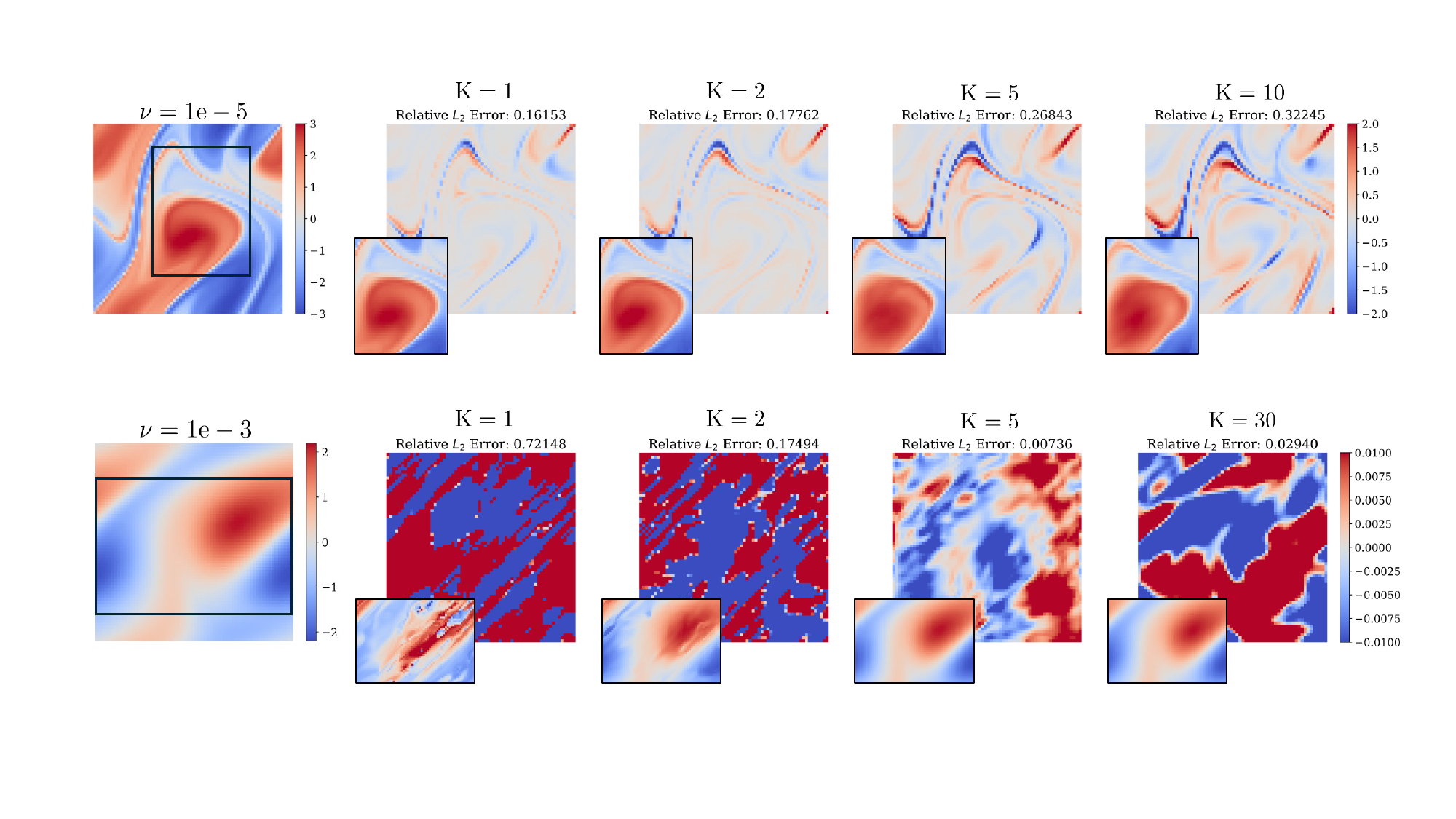}
    \caption{\textbf{Performance of TB-Transolver on the Navier-Stokes Benchmark}: For \textbf{low viscosity} (top), as bundle width ${\rm K}$ increases toward ${\rm T_{out}}$, TB-Transolver achieves a speedup of up to $3.4\times$ while relative $\rm L_2$ error increases by $115\%$ compared to autoregressive rollouts at ${\rm K} = 1$. On the other hand, for \textbf{high viscosity} (bottom), as bundle width ${\rm K}$ increases toward ${\rm T_{out}}$, TB-Transolver achieves a speedup of up to $12\times$ while relative $\rm L_2$ error decreases by $99\%$ compared to autoregressive rollouts at ${\rm K} = 1$.} 
    \label{fig:navier_stokes}
\end{figure}

\begin{table}[H]
\centering
\begin{tabular}{ccccccc}
\toprule
Bundle Width ($\rm K$) & Mean Error & Sec/Epoch  & Total Time (500 Ep.) & Peak VRAM & $\rm C(K)$ & Speed-up\\
\midrule
1  & $0.09618^{[1]}$ & 79.7 & 12.0 h & 7.65 GB & 1.0 & $\rm Baseline$\\
2  & $0.10638^{[1]}$ & 47.4 & 6.5 h & 3.94 GB & 0.935 & $1.7\times$\\
5  & $0.16571^{[1]}$ & 31.6 & 4.2 h & 1.72 GB & 0.696 & $2.5\times$\\
10 & $0.20670^{[1]}$ & 23.4 & 2.7 h & 0.98 GB & 0.430 & $3.4\times$\\
\bottomrule
\end{tabular}
\caption{\textbf{Navier-Stokes, Low Viscosity Benchmark ($\rm T_{out}=10$)}: Accuracy and computational efficiency for varying bundle widths. ${\rm C(K)}$ follows the trend illustrated in Figure~\ref{fig:temporal_correlation}. The model is trained using autoregressive time rollouts. The model comprises $11.2$ million parameters.}
\label{tab:navier_stokes}
\end{table}

Notably, the plasticity and low-viscosity Navier--Stokes benchmarks exhibit similar decaying trends in $\rm C(K)$ as the bundle width $\rm K$ increases, as shown in Figure~\ref{fig:temporal_correlation}. This decay indicates increasing decorrelation between the first and last snapshots within a temporal bundle. Consistent with this observation, the prediction accuracy degrades as $\rm K$ increases for both benchmarks, as reported in Tables~\ref{tab:plasticity}--\ref{tab:navier_stokes}, due to increased decoherence within the bundle. For plasticity, evaluated using time-conditioned rollouts, temporal bundling achieves up to $29\times$ speed-up, while the low-viscosity Navier--Stokes benchmark, evaluated using autoregressive rollouts, achieves up to $3.4\times$ speed-up. However, at $\rm K = T_{out}$, the mean error increases by up to $2.25\times$ and $2.24\times$, respectively, relative to the $\rm K = 1$ baseline. \emph{This suggests that for any \textbf{decaying trend of $\rm C(K)$} as observed for plasticity (Table \ref{tab:plasticity}) and Navier-Stokes with low viscosity (Table \ref{tab:navier_stokes}) benchmarks, an optimal bundle width $\rm K^*$ is typically when $\rm K = 2$ where we attain up to $2\times$ speed-up with minimal increase in the mean error.} 

\begin{figure}[H]
    \centering
    \includegraphics[width=1\linewidth, trim={2cm 5.3cm 2cm 3.5cm}, clip]{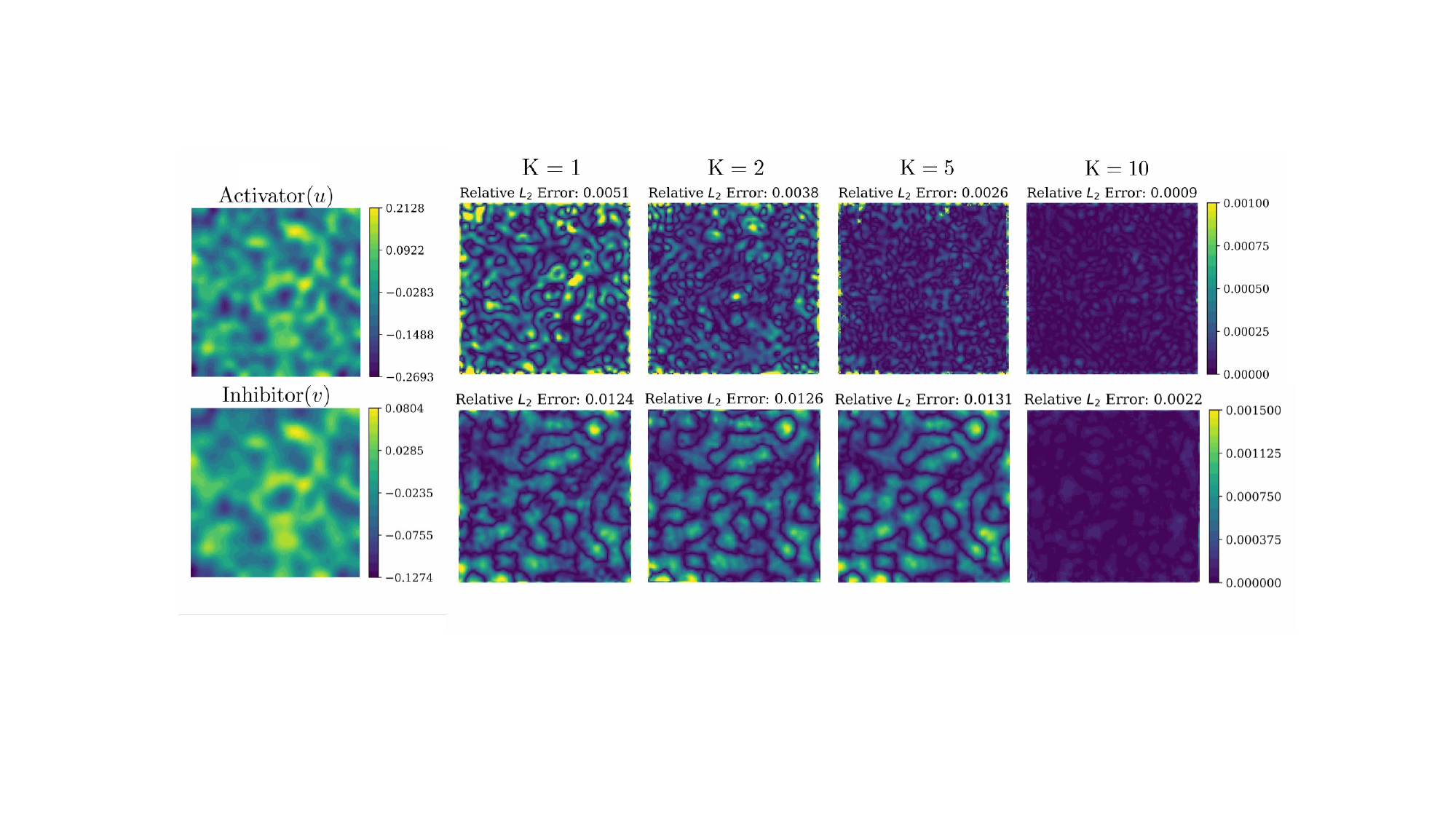}
    \caption{\textbf{Performance of TB-Transolver on the Reaction--Diffusion Benchmark}: As bundle width ${\rm K}$ increases toward ${\rm T_{out}}$, TB-Transolver achieves a speedup of up to $10\times$ alongside substantial reduction in relative $\rm L_2$ error by $76.9\%$ compared to autoregressive rollouts at ${\rm K} = 1$.}
    \label{fig:reaction_diffusion}
\end{figure}

\begin{table}[H]
\centering
\begin{tabular}{ccccccc}
\toprule
Bundle Width ($\rm K$) & Mean Error & Sec/Epoch & Total Time (500 Ep.) & Peak VRAM & $\rm C(K)$ & Speed-up\\
\midrule
1  & $0.00489^{[1]}$ & 216.00 & 30.0 h & 29.83 GB & 1.0 & $\rm Baseline$\\
2  & $0.00438^{[1]}$ & 109.15 & 15.2 h & 15.05 GB & 0.999 & $2\times$\\
5  & $0.00394^{[2]}$ & 44.20  & 6.1 h  & 6.18 GB & 0.998 & $5\times$\\
10 & $0.00113^{[2]}$ & 22.60  & 3.1 h  & 3.22 GB & 0.994 & $10\times$\\
\bottomrule
\end{tabular}
\caption{\textbf{Reaction-Diffusion Benchmark ($\rm T_{\text{out}}=10$)}: Accuracy and computational efficiency for varying bundle widths. ${\rm C(K)}$ follows the trend illustrated in Figure~\ref{fig:temporal_correlation}. The model is trained using autoregressive time rollouts. The model comprises $11.2$ million parameters.}
\label{tab:reaction_diffusion}
\end{table}

In contrast, the reaction--diffusion and global sea surface temperature (SST) forecasting benchmarks exhibit non-decaying, approximately flat trends in $\rm C(K)$ as the bundle width $\rm K$ increases, as shown in Figure~\ref{fig:temporal_correlation}. This behavior indicates that the dynamics remain coherent across the temporal bundle, even at larger bundle widths. Consistent with this observation, prediction accuracy is maintained or improves as $\rm K$ increases for the reaction--diffusion and SST benchmarks, as reported in Tables~\ref{tab:reaction_diffusion}--\ref{tab:sst}. The reaction--diffusion benchmark, evaluated using autoregressive rollouts, achieves up to $10\times$ speed-up, while the SST benchmark achieves up to $7.5\times$ speed-up. For both benchmarks, the lowest mean error is obtained at $\rm K = T_{out}$ relative to the baseline with $\rm K = 1$. \emph{These results suggest that, for benchmarks exhibiting a \textbf{non-decaying or flat trend in $\rm C(K)$}, such as reaction--diffusion (Table~\ref{tab:reaction_diffusion}) and SST forecasting (Table~\ref{tab:sst}), an optimal bundle width $\rm K^*$ is typically when $\rm K \rightarrow T_{out}$, yielding the greatest computational efficiency alongside the best test accuracy.}

\begin{figure}[H]
    \centering    \includegraphics[width=\linewidth, trim={2cm 7.5cm 1cm 6cm}, clip]{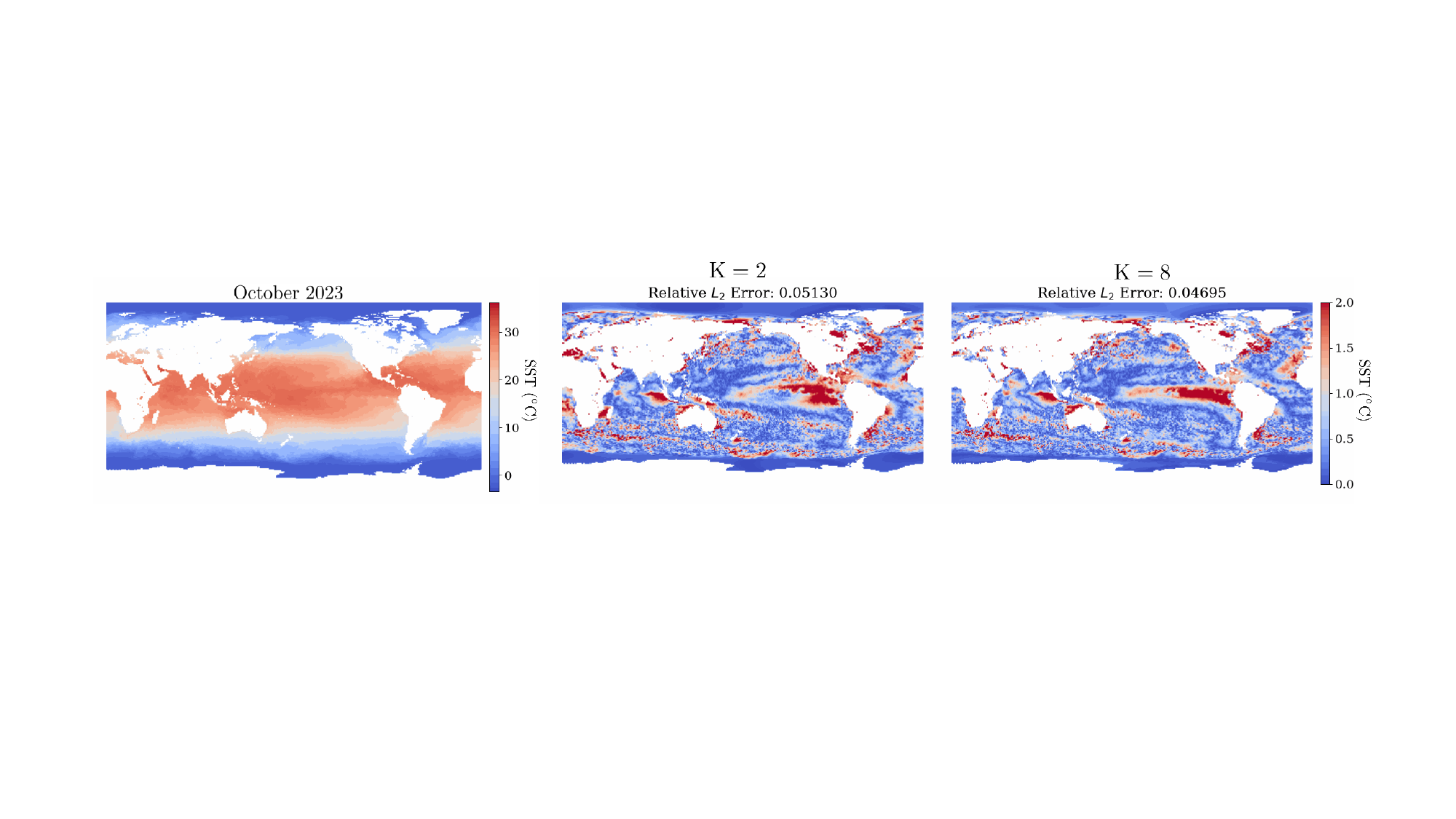}
    \caption{\textbf{Performance of TB-Transolver on the SST Benchmark}: As bundle width ${\rm K}$ increases toward ${\rm T_{out}}$, TB-Transolver achieves a speedup of up to $7.5\times$ alongside reduction in relative $\rm L_2$ error by $7.9\%$ compared to time-bundled rollouts at ${\rm K} = 2$. Note that ${\rm K} = 1$ exhausts available GPU memory for a batch size of $8$ at \texttt{float32} precision. The test dataset spans years 2021–2025, with additional evaluation results provided in Appendix \ref{appx:sst}.}
    \label{fig:sst}
\end{figure}

\begin{table}[H]
\centering
\begin{tabular}{ccccccc}
\toprule
Bundle Width ($\rm K$) & Mean Error & Sec/Epoch & Total Time (500 Ep.) & Peak VRAM & $\rm C(K)$ & Speed-up\\
\midrule
1  & \multicolumn{3}{c}{ ------------ Out Of Memory (OOM) ------------ } & 94.86 GB* & 1.0 & ---\\
2  & $0.08000^{[3]}$  & 4.22 & 35.2 min & 54.72 GB & 0.993 & $\rm Baseline$\\
4  & $0.07550^{[3]}$  & 2.12 & 17.7 min & 27.53 GB & 0.954 & $1.9\times$\\
8  & $0.07370^{[3]}$  & 1.07 & 8.9 min  & 13.93 GB & 0.994 & $3.9\times$\\
\bottomrule
\multicolumn{5}{l}{\footnotesize *Maximum available GPU memory}
\end{tabular}
\caption{\textbf{Sea Surface Temperature (SST) Benchmark ($\rm T_{\text{out}}=8$ months)}: Accuracy and computational efficiency for varying bundle widths. ${\rm C(K)}$ follows the trend illustrated in Figure~\ref{fig:temporal_correlation}. The model is trained using autoregressive time rollouts. The model comprises 710k parameters.}
\label{tab:sst}
\end{table}

The benchmarks discussed above can be categorized as involving short to medium-horizon rollouts. In contrast, we consider high-viscosity Navier--Stokes as a long-horizon benchmark, predicting 30 future time snapshots given 20 historical input snapshots. Unlike the low-viscosity regime, the high-viscosity regime exhibits more temporally coherent, periodically oscillating flow structures, which may favor temporal bundling and help explain the oscillations in $\rm C(K)$ at larger bundle widths, as observed in Figure~\ref{fig:temporal_correlation}. Furthermore, we observe a difference of approximately $\rm \Delta C(K) \approx 0.1$ between the low and high-viscosity Navier--Stokes benchmarks. Taken together, these observations suggest that the high-viscosity regime represents an intermediate case between the decaying and flat trends in $\rm C(K)$ observed for the other benchmarks.

\begin{table}[H]
\centering
\begin{tabular}{ccccccc}
\toprule
Bundle Width ($\rm K$) & Mean Error & Sec/Epoch & Total Time (500 Ep.) & Peak VRAM & $\rm C(K)$ & Speed-up\\
\midrule
1  & $0.54380$ & 243.83 & 33.9 h & 22.49 GB & 1.0 & $\rm Baseline$\\
2  & $0.08000$ & 123.60 & 17.2 h & 11.37 GB & 0.984 & $2\times$\\
5  & $0.00530$ & 54.17  & 7.5 h  & 4.69 GB & 0.819 & $4.5\times$\\
10 & $0.00687$ & 38.43  & 5.3 h  & 2.46 GB & 0.505 & $6.4\times$\\
15 & $0.00603$ & 33.20  & 4.6 h  & 1.72 GB & 0.665 & $7.3\times$\\
30 & $0.00714$ & 20.16  & 2.8 h  & 0.98 GB & 0.454 & $12\times$\\
\bottomrule
\end{tabular}
\caption{\textbf{Navier-Stokes, High-Viscosity Benchmark ($\rm T_{\text{out}}=30$)}: Accuracy and computational efficiency for varying bundle widths. ${\rm C(K)}$ follows the trend illustrated in Figure~\ref{fig:temporal_correlation}. The model is trained using autoregressive time rollouts. The model comprises 11.2 million parameters.}
\label{tab:high_viscosity_ns}
\end{table}

As the bundle width $\rm K$ increases, we observe a monotonically decreasing trend in training time but a non-monotonic trend in mean error, despite the decaying trend in $\rm C(K)$. As shown in Table~\ref{tab:high_viscosity_ns}, the optimal bundle width is $\rm K^* = 5$. We can consider $\rm K < K^*$ as a less favorable region, as it incurs substantially higher training time and mean error. In contrast, $\rm K > K^*$ substantially reduces training time, although the mean error increases, while maintaining the same order of magnitude as that obtained at the optimum. \emph{This suggests that for decaying trend of $\rm C(K)$ with long horizon rollouts, as observed for Navier-Stokes with high viscosity (Table \ref{tab:high_viscosity_ns}) benchmark, an optimal bundle width $\rm K^* = 5$ exists in the range, $\rm 0 < K^* < T_{out}$ that attains up to $4.5\times$ speed-up with maximum test accuracy.}

\begin{remark}[\textbf{Physical Interpretation of the Influence of Temporal Bundling on the Navier--Stokes Benchmark}]
For the \textbf{low-viscosity benchmark}, the vorticity field in the highlighted inset of Figure~\ref{fig:navier_stokes} shows that the flow becomes increasingly diffusive as ${\rm K}$ increases. Physically, this behavior suggests that wider temporal bundles introduce artificial numerical dissipation, which acts in addition to the physical viscosity ($\nu = 10^{-5} + \nu_{\text{artificial}}$) and enhances diffusion relative to the baseline due to decoherence within the bundle. Although a similar qualitative trend is observed for the \textbf{high-viscosity benchmark}, the contribution of this artificial damping to the relative ${\rm L_2}$ error is substantially smaller than that of the distribution shift induced by long-horizon autoregressive rollouts. Consequently, distribution shift dominates the error in the high-viscosity regime, resulting in a substantially larger error at ${\rm K} = 1$ than in the low-viscosity benchmark, which achieves its lowest overall error at this bundle width.
\end{remark}

\begin{remark}[\textbf{Memory Efficiency}] 
All benchmark results discussed in this section pertain to single-GPU training of the temporally bundled Transolver under a fixed VRAM budget of 96 GB for RTX PRO 6000 Blackwell. Across all domains, we observe a drastic reduction in VRAM consumption for autoregressive time rollouts as the bundle width increases. Conversely, for time-conditioned rollouts, VRAM consumption remains invariant across bundle widths (as shown in Table \ref{tab:plasticity} for plasticity) because the model does not accumulate extended computational graphs prior to backpropagation at the end of the predicted time horizon. Thus, while time-conditioned rollouts are insensitive to bundle width changes, autoregressive rollouts exhibit substantial VRAM savings as $\rm K \to T_{out}$. Although secondary factors such as batch size also influence memory usage, we note that for the SST benchmark with a batch size of 8, the combined memory footprint of the dataset and model parameters (in \texttt{float32}) exceeds the GPU capacity, resulting in an out-of-memory error as shown in Table~\ref{tab:sst}.
\end{remark}

\subsection{Distributed Data Parallelism at Optimal Bundle Width}

In this section, training is extended to multiple GPUs using Distributed Data Parallelism (DDP). Specifically, we leverage multiple NVIDIA RTX PRO 6000 Blackwell GPUs to scale the training of TB-Transolver across time-dependent PDE benchmarks. In addition to identifying an optimal bundle width $\rm K^*$ that balances prediction accuracy and computational efficiency, we evaluate the additional speedups achieved through multi-GPU scaling.

\begin{table}[H]
\centering
\begin{tabular}{lccccc}
\toprule
\multirow{2}{*}{Benchmark} & Optimal Bundle & 1-GPU & 2-GPU DDP & \multirow{2}{*}{Speed-up} & \multirow{2}{*}{$\%$ Scaling Efficiency} \\
& Width ($\rm K^*$) & (Sec/Epoch) & (Sec/Epoch) & & \\
\midrule
Plasticity             & 2  & 62.89 & 32.58 & 1.93$\times$ & 96.5 \\
Low Viscosity           & 2  & 47.40 & 25.09 & 1.89$\times$ & 94.5 \\
High Viscosity          & 5  & 54.17 & 28.23 & 1.92$\times$ & 96.0 \\
Reaction-Diffusion     & 10 & 22.60 & 14.23 & 1.59$\times$ & 79.5 \\
SST   & 8  &  1.07 &  0.57 & 1.88$\times$ & 94.0 \\
\bottomrule
\end{tabular}
\caption{\textbf{Distributed Data Parallel (DDP) Scaling}: Scaling efficiency up to 2 GPUs evaluated across diverse physical regimes using optimal temporal bundle widths ($\rm K^*$). Near-linear scaling is achieved on compute-heavy domains where Physics-Attention overhead dominates communication latency.}
\label{tab:ddp_scaling_overview}
\end{table}

\noindent For multi-GPU training, the GPUs are distributed across compute nodes, with two GPUs per node. Each compute node is equipped with two NVIDIA RTX PRO 6000 Blackwell GPUs. The GPUs support PCIe Gen5 x16 connectivity but do not support NVLink. For instance, in the 8-GPU configuration, data is distributed across GPUs spanning four compute nodes, with inter-GPU communication within the node facilitated via PCIe rather than the higher-bandwidth NVLink. Although NVLink is typically preferred for high-bandwidth, low-latency peer-to-peer GPU communication, the PCIe-based intra node connectivity and inter-node communication infrastructure are sufficient for the current workload, with an average measured communication speed of $700\text{ MB/s}$. This efficiency is primarily attributed to single-precision (\texttt{float32}) training paired with modest batch sizes. Nevertheless, GPU architectures equipped with NVLink may be advantageous for memory-bound scaling scenarios involving larger models, higher-precision training, larger batch sizes, or larger datasets.

It is worth emphasizing that the speedups achieved through multi-GPU training are orthogonal to those achieved through temporal bundling. Specifically, temporal bundling accelerates training along the time dimension by shortening computational graphs, whereas multi-GPU training distributes computation across GPUs to improve scalability. Regardless of the number of GPUs, time-conditioned and autoregressive rollouts process time steps sequentially by default. Thus, specialized strategies such as temporal bundling remain essential for mitigating computational bottlenecks along the time axis.

\begin{figure}[H]
    \centering
    
    \subfloat[]{
        \includegraphics[width=0.75\linewidth, trim={0 1cm 0 0}, clip]{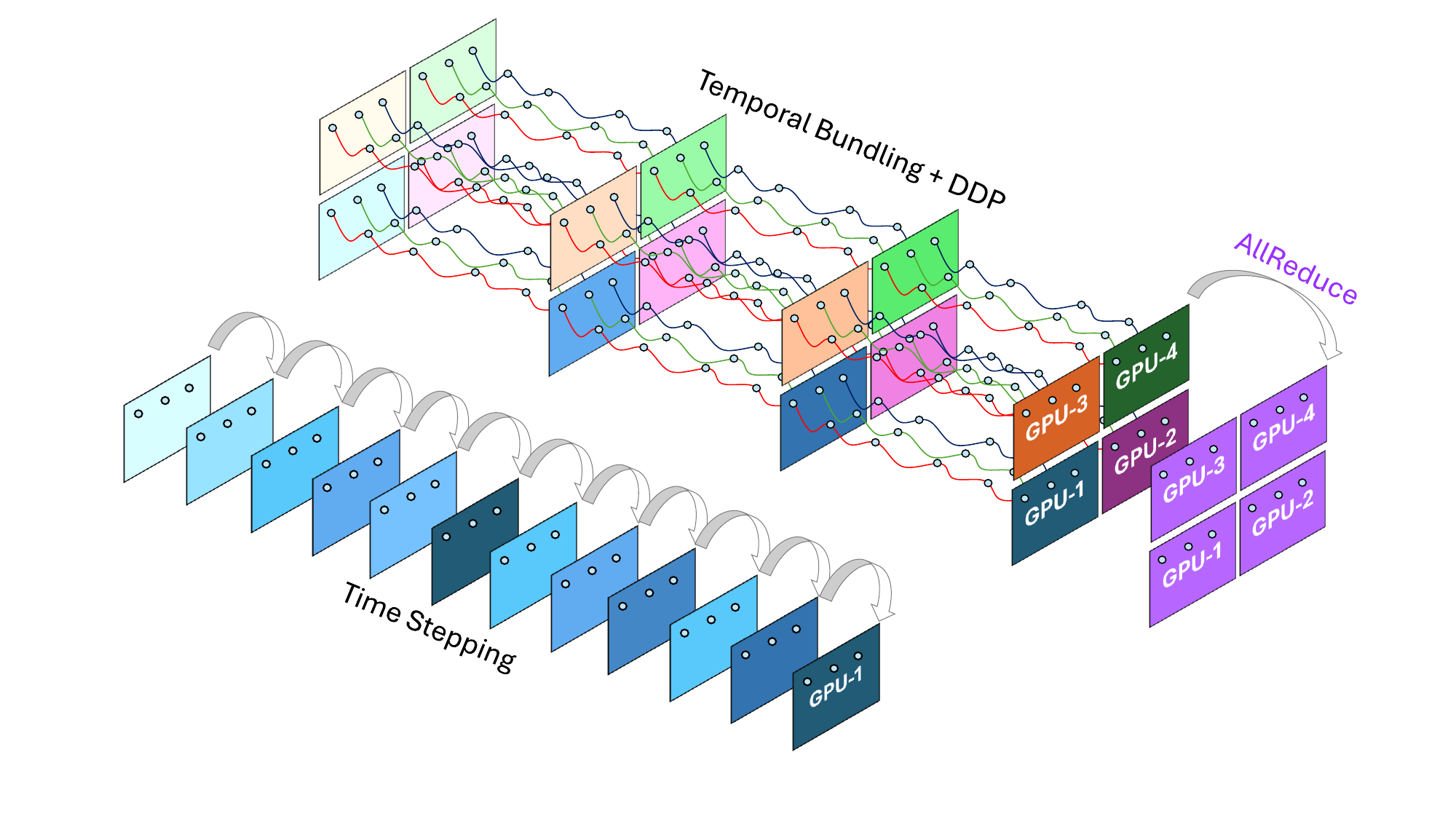}
        \label{fig:accuracy_motivation}
    }
    \vspace{1.5em}
    \subfloat[]{
        \includegraphics[width=0.85\linewidth]{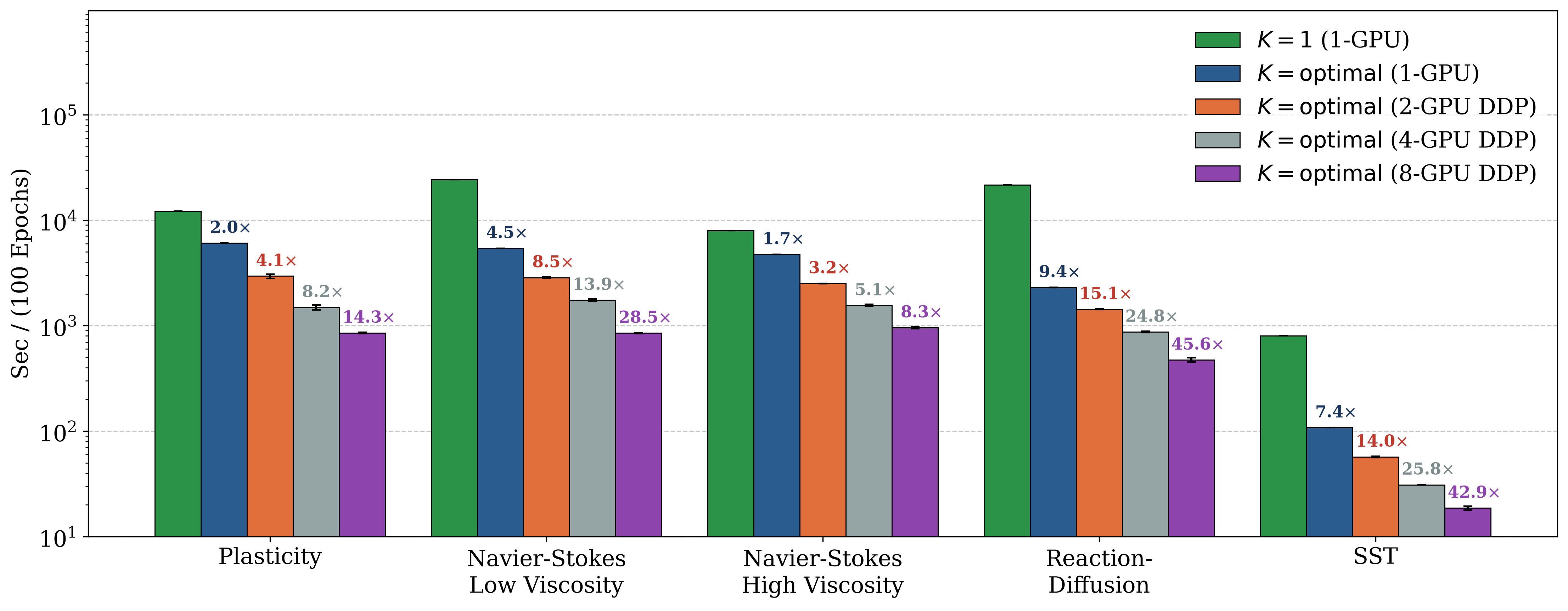}
        \label{fig:ddp_speedup_main}
    }
    \caption{\textbf{Multi-GPU DDP Scaling with Temporal Bundling}: (a) Illustration comparing regular time stepping on a single GPU and multi-GPU training with temporal bundling for a 4-GPU configuration. Gradients are synchronized across GPUs using the AllReduce operation during backpropagation, before the optimizer update. (b) Multi-GPU scaling up to 8 GPUs across diverse physical domains using optimal bundle width $\rm K^*$ with compute load dominating communication latency.}
    \label{fig:ddp_scaling}
\end{figure}

\noindent Table~\ref{tab:ddp_scaling_overview} summarizes the scaling efficiency and speedups achieved across the time-dependent benchmarks under a 2-GPU configuration, relative to single-GPU training at the optimal bundle width $\rm K^*$. With the exception of the reaction--diffusion benchmark, all evaluated domains attain an average scaling efficiency of $95\%$ under the 2-GPU configuration. We further scale training to an 8-GPU configuration, with the resulting performance trends illustrated in Figure~\ref{fig:ddp_scaling}. Key observations from scaling TB-Transolver at $\rm K^*$ include:
\begin{itemize}
    \item As shown in Figure~\ref{fig:ddp_scaling}, combining temporal bundling with multi-GPU training yields aggregate speedups of up to $45.6\times$ across the evaluated benchmarks. These gains arise from accelerating computation along the time dimension through temporal bundling and distributing computation across multiple GPUs.
    \item The reaction--diffusion and SST benchmarks achieve aggregate speedups exceeding $40\times$ with 8 GPUs, supported by their large optimal bundle widths ($\rm K^* = 10$ and $\rm K^* = 8$, respectively). These results highlight the substantial contribution of temporal bundling at larger bundle widths to the overall computational gains, complemented by multi-GPU scaling.
    \item In contrast, the long-horizon autoregressive high-viscosity Navier--Stokes benchmark achieves a maximum aggregate speedup of $8.3\times$ with 8 GPUs, the lowest among the evaluated benchmarks. This comparatively modest gain may reflect the computational demands of its longer input and output time sequences, despite the use of an optimal bundle width of $\rm K^* = 5$.
    \item Across the five benchmarks, 8-GPU distributed training achieves average GPU utilization of approximately $81$--$94\%$, indicating sustained and relatively balanced GPU activity across the workloads. Streaming Multiprocessor (SM) utilization ranges from $30$--$60\%$, while DRAM bandwidth utilization varies from $10$--$50\%$, reflecting differences in computational and memory-access demands across physical systems. The SST benchmark exhibits the highest GPU memory consumption (approximately $16.5$ GB per GPU), primarily due to its larger batch size of $8$.
\end{itemize}

\noindent Temporal bundling and multi-GPU scaling are complementary mechanisms for improving the computational efficiency of TB-Transolver, but their relative contributions vary across benchmarks. Our results suggest that this variation is associated with the trend in $\rm C(K)$ across the time-dependent datasets. For benchmarks where $\rm C(K)$ decays, as illustrated in Figure~\ref{fig:temporal_correlation}, the accuracy constraints on bundle width can limit the speedup obtainable through temporal bundling, making multi-GPU scaling an important additional source of computational gains. Conversely, when $\rm C(K)$ remains approximately flat, larger bundle widths can be used while maintaining prediction accuracy, allowing temporal bundling to contribute more substantially to the overall speedup.

\section{Discussion \texorpdfstring{$\&$}{\&} Conclusions}

\begin{table}[H]
\centering
\small
\begin{tabular}{l c c c}
\toprule
\textbf{Benchmark} & \textbf{$\rm K^*$} & \textbf{$\rm C(K^*)$} & \textbf{Speed-up}\\
\midrule
Plasticity & 2 & 0.894 & $2\times$\\
Navier-Stokes (Low Viscosity) & 2 & 0.935 & $1.7\times$\\
Navier-Stokes (High Viscosity) & 5 & 0.819 & $4.5\times$\\
Reaction-Diffusion & 10 & 0.994 & $10\times$\\
Sea Surface Temperature (SST) & 8 & 0.994 & $3.9\times$ \\
\bottomrule
\end{tabular}
\caption{\textbf{Optimal bundle widths and their accuracy-efficiency trade-offs}:
Following the selection criterion in Equation \eqref{eq:c_k_threshold}: $\rm K^* = \max \left\{ K \;:\; C(K) \gtrsim 0.8 \right\}$, the selected optimal bundle widths ($\rm K^*$) provide computational speed-ups while maintaining favorable predictive accuracy across the benchmarks.}
\label{tab:optimal_k_benchmark}
\end{table}

The Transolver architecture has been widely adopted as a neural operator framework due to its efficient Physics-Attention mechanism, which enables the handling of large geometric meshes in physical space. However, it often struggles with scalability in time-dependent physical problems, restricting its utility for complex dynamical systems. Specifically, much of this inefficiency stems from sequential time-stepping via time-conditioned or autoregressive training, causing the architecture to fall short on temporal dynamics despite the near-linear computational complexity of its spatial attention mechanism. To address these limitations, this work proposes the TB-Transolver as an efficient framework for solving time-dependent problems. Furthermore, we introduce spatiotemporal coherence, $\rm C(K)$, as a metric to quantify temporal coherence and determine the optimal bundle width $\rm K^*$ for the temporally bundled Transolver. As illustrated in Table \ref{tab:optimal_k_benchmark}, all benchmarks satisfy the criterion in Equation \eqref{eq:c_k_threshold}, achieving computational speed-ups while maintaining a favorable accuracy-efficiency trade-off at the optimal bundle width, $\rm K^*$. It is observed that the source of error for temporal bundling with an arbitrary bundle width $\rm K$ can largely vary depending on the trend of $\rm C(K)$ and this is illustrated in Table \ref{tab:benchmark_analysis_source_error}. For instance, as $\rm C(K)$ decays the error contribution is dominated by decoherence for $\rm K \rm > 1$ with an exception for the high-viscosity benchmark. Consequently, when $\rm C(K)$ plateaus, the contribution is only from accumulation for $\rm 1 < K < T_{out}$.

\begin{table}[H]
\centering
\small
\begin{tabular}{l c cc}
\toprule
\multirow{2}{*}{\textbf{Benchmark}} & \multirow{2}{*}{\textbf{$\rm C(K)$}} & \multicolumn{2}{c}{\textbf{Source of error dominated by}} \\
\cmidrule(lr){3-4}
 & & $\rm \varepsilon_{accumulation}$ & $\rm \varepsilon_{decoherence}$ \\
\midrule
Plasticity & Decays & - & \cmark \\
Navier-Stokes (Low Viscosity) & Decays & \xmark & \cmark \\
Navier-Stokes (High Viscosity) & Decays and Oscillates & \cmark & \xmark \\
Reaction-Diffusion & Flat & \cmark & - \\
Sea Surface Temperature (SST) & Flat & \cmark & - \\
\bottomrule
\end{tabular}
\caption{\textbf{Sources of error in temporal bundling}: Error is governed by either accumulation or decoherence for $1 < {\rm K} < {\rm T_{out}}$. $\rm \varepsilon_{decoherence}$ can be neglected in regimes where ${\rm C(K)}$ remains flat. Plasticity benchmark has no contribution from $\rm \varepsilon_{accumulation}$ since it is trained by time-conditioned rollouts. Note that this is a qualitative interpretation of the dominant error mechanisms.}
\label{tab:benchmark_analysis_source_error}
\end{table}

\noindent Although temporal bundling yields exceptional gains in computational efficiency, it often struggles to maintain accuracy at large bundle widths due to decoherence for certain time-dependent benchmarks. In that context, the potential of temporal bundling is not fully leveraged for such benchmarks since they are forced to work with lower bundle widths to balance computational efficiency and accuracy. To mitigate this limitation, we integrate temporal bundling with multi-GPU training, securing cumulative speedups particularly in regimes where accuracy constraints restrict the use of larger bundle widths. Consequently, the relative dominance of temporal bundling versus multi-GPU scaling in driving computational efficiency varies significantly across domains depending on the trend of $\rm C(K)$, as summarized in Table~\ref{tab:benchmark_analysis_efficiency}. To this end, a monotonically decaying ${\rm C(K)}$ results in significant internal decoherence at large bundle widths, restricting optimal performance to smaller ${\rm K}$ values and thereby diminishing the computational gains achievable via temporal bundling. In such regimes, hardware-level GPU scaling serves as the primary driver of computational efficiency. Conversely, in systems where the trend of ${\rm C(K)}$ plateaus, high structural coherence is preserved even at larger bundle widths, allowing temporal bundling to yield substantial efficiency gains that surpass those provided by GPU scaling alone. This is further supported by the empirical evidence illustrated in Figure \ref{fig:ddp_scaling}.

\begin{table}[H]
\centering
\small
\begin{tabular}{l c cc}
\toprule
\multirow{2}{*}{\textbf{Benchmark}} & \multirow{2}{*}{\textbf{$\rm C(K)$}} & \multicolumn{2}{c}{\textbf{Computational efficiency dominated by}} \\
\cmidrule(lr){3-4}
 & & \textbf{Temporal bundling} & \textbf{GPU scaling} \\
\midrule
Plasticity & Decays & \xmark & \cmark \\
Navier-Stokes (Low Viscosity) & Decays & \xmark & \cmark \\
Navier-Stokes (High Viscosity) & Decays and Oscillates & \omark & \omark \\
Reaction-Diffusion & Flat & \cmark & \xmark \\
Sea Surface Temperature (SST) & Flat & \cmark & \xmark \\
\bottomrule
\end{tabular}
\caption{\textbf{Computational efficiency across domains for temporal bundling}: Computational efficiency across distinct physical domains, governed by the combined effects of temporal bundling and multi-GPU scaling.}
\label{tab:benchmark_analysis_efficiency}
\end{table}

\noindent In summary, the present work addresses the primary cause of computational inefficiency in the Transolver architecture during time-dependent PDE rollouts. Beyond eliminating the scalability bottleneck, we introduce a predictive framework based on spatiotemporal coherence ${\rm C(K)}$ to systematically determine the optimal bundle width ${\rm K^*}$ using a priori analysis of the training data alone, bypassing the need for computationally prohibitive grid or random searches. Importantly, the proposed coherence diagnostic is formulated independently
of the specific neural operator architecture and offers a data-driven
approach to assessing temporal rollout efficiency. Its applicability
across broader classes of neural operators, however, remains to be
established through future investigations.


Extending this validation to demonstrate the broader generalizability of the proposed framework across alternative neural operator architectures prominent in the literature represents a promising direction for future work. Finally, the proposed strategies for TB-Transolver can be seamlessly extended to (3+1)D industry-scale problems, such as crash dynamics \cite{nabian2026automotive} or automotive aerodynamics \cite{wu2024transolver}, even in resource-constrained environments. Furthermore, when combined with high-performance computing hardware such as GPUs, the framework attains massive aggregated speedups while preserving predictive accuracy and strictly respecting the temporal causality of time-stepping rollouts at optimal bundle widths. Another promising direction involves extending TB-Transolver to discrete physical systems such as granular metamaterials, departing from the continuous domains predominantly investigated in existing literature. We believe this extension will require tailored architectural modifications to effectively accommodate physical discreteness, representing a compelling direction for future research.

\appendix
\section{Training Configuration}
\label{appx:training_config}

All Transolver and TB-Transolver models were trained using the same configuration for each time-dependent benchmark. The AdamW optimizer was used with a weight decay of \(10^{-5}\) and an initial learning rate of \(10^{-3}\). The learning rate was scheduled using PyTorch's OneCycleLR scheduler. All models were trained for 500 epochs.

\begin{table}[H]
\centering
\small
\setlength{\tabcolsep}{4pt}
\begin{tabular}{lccccc}
\hline
\textbf{Parameter} 
& \textbf{Plasticity} 
& \textbf{NS (Low $\nu$)} 
& \textbf{Reaction--Diffusion} 
& \textbf{SST} 
& \textbf{NS (High $\nu$)} \\
\hline
Spatial domain 
& 2D structured 
& 2D structured 
& 2D structured 
& 2D irregular 
& 2D structured \\

Hidden dimension 
& 128 & 256 & 256 & 128 & 256 \\

Number of layers 
& 8 & 8 & 8 & 8 & 8 \\

Number of attention heads 
& 8 & 8 & 8 & 8 & 8 \\

Number of learned slices 
& 64 & 32 & 32 & 32 & 32 \\

Unified positional encoding 
& Disabled & Enabled & Enabled & Disabled & Enabled \\

Input time steps ($\rm T_{\rm in}$) 
& -- & 10 & 10 & 4 months & 20 \\

Output time steps ($\rm T_{\rm out}$) 
& 20 & 10 & 10 & 8 months & 30 \\

Rollout strategy 
& Time-conditioned & Autoregressive & Autoregressive & Autoregressive & Autoregressive \\

Teacher forcing 
& Disabled & Enabled & Enabled & Enabled & Enabled \\

Training samples 
& 900 & 1,000 & 900 & 41 & 1,000 \\

Test samples 
& 80 & 200 & 100 & 5 years & 200 \\

Batch size per GPU 
& 2 & 2 & 2 & 8 & 2 \\

Spatial resolution 
& $101 \times 31$ & $64 \times 64$ & $64 \times 64$ & 43,049 points & $64 \times 64$ \\
\hline
\end{tabular}
\caption{Transolver architecture and training configurations across the plasticity, low- and high-viscosity Navier--Stokes, reaction--diffusion, and sea surface temperature (SST) benchmarks.}
\label{tab:transolver_benchmark_config}
\end{table}

\section{Sea Surface Temperature}
\label{appx:sst}
\begin{figure}[H]
    \centering    \includegraphics[width=\linewidth, trim={2cm 4.5cm 1cm 3.9cm}, clip]{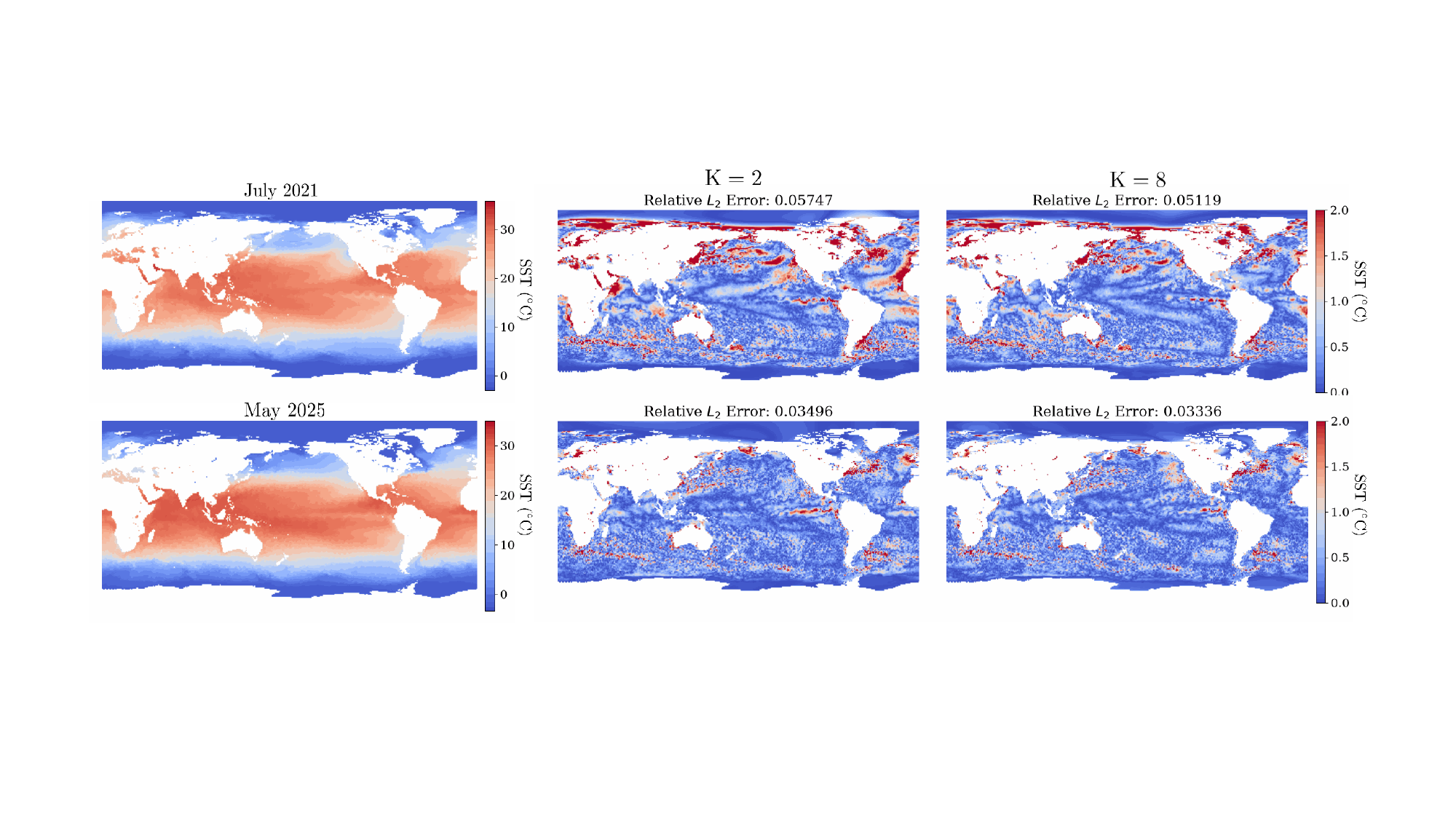}
    \caption{\textbf{Performance of TB-Transolver on the SST Benchmark}: As bundle width ${\rm K}$ increases toward ${\rm T_{out}}$, TB-Transolver achieves a speedup of up to $7.5\times$ alongside reduction in relative $\rm L_2$ error by $7.9\%$ compared to time-bundled rollouts at ${\rm K} = 2$. Note that ${\rm K} = 1$ exhausts available GPU memory for a batch size of $8$ at \texttt{float32} precision.}
    \label{fig:sst_2}
\end{figure}

\bibliographystyle{elsarticle-num} 
\bibliography{reference}
\end{document}